\newif\ifdraft

\documentclass[final]{iasart}			

\usepackage[pdftex]{graphicx}
\graphicspath{{./fig/}}
\DeclareGraphicsExtensions{.pdf,.jpg,.png}

\usepackage[utf8]{inputenc}
\usepackage[T1]{fontenc}

\usepackage{xcolor}
\colorlet{BLUE}{blue} 
\colorlet{BLACK}{black}

\usepackage{natbib}
\usepackage[hidelinks]{hyperref} 
\usepackage{doi}

\usepackage{textcomp} 

\usepackage{url}

\usepackage{amsthm}

\usepackage{amsfonts}
\usepackage{amsmath,bm}

\usepackage{mathrsfs}
\usepackage{eucal}
\usepackage{scalerel}

\usepackage{array} 

\usepackage[caption=false,font=footnotesize]{subfig}

\usepackage{stfloats}

\usepackage{paralist}

\usepackage[all]{nowidow}

\newcommand\scale[2]{\vstretch{#1}{\hstretch{#1}{#2}}}
	
\newcommand{\LP}{
		\mathbin{\mathpalette\LIPcls+}
	}
\newcommand{\LM}{
		\mathbin{\mathpalette\LIPcls-}
		}
\newcommand{\LT}{
		\mathbin{\mathpalette\LIPcls\times}
	}

\newcommand{\LIPcls}[2]{%
		\ooalign{$#1\bigtriangleup$\crcr  \hidewidth\raisefix{#1}\hbox{$#1\scale{0.45}{\bm{#2}}$}\hidewidth}}

\def\raisefix#1{%
  \ifx#1\displaystyle
    \raise.14em
  \else
    \ifx#1\textstyle
      \raise.14em
    \else
      \ifx#1\scriptstyle
        \raise.112em
      \else
        \raise.0933em
      \fi
    \fi
  \fi
}

\newcommand{\LIPplus}{\LP}
\newcommand{\LIPminus}{\LM}

\makeatletter
\DeclareRobustCommand\bigop[1]{%
  \mathop{\vphantom{\sum}\mathpalette\bigop@{#1}}\slimits@
}
\newcommand{\bigop@}[2]{%
  \vcenter{%
    \sbox\z@{$#1\sum$}%
    \hbox{\resizebox{\ifx#1\displaystyle1.1\fi\dimexpr\ht\z@+\dp\z@}{!}{$\m@th#2$}}%
  }%
}
\makeatother

\newcommand{\Real}{\mathbb R}
\newcommand{\Realb}{\overline{\Real}}
\newcommand{\Tcurv}{\mathcal{T}}
\newcommand{\Tcurvb}{\overline{\Tcurv}}

\newcommand{\la}{\lambda}

\newcommand{\I}{\mathcal{I}}
\newcommand{\Ib}{\overline{\I}}

\newcommand{\Fcurv}{\mathcal{F}}
\newcommand{\Fcurvb}{\overline{\Fcurv}}

\newcommand{\Lscr}{\mathscr{L}}

\newtheorem{proposition}{Proposition}
\newtheorem{corollary}{Corollary}
\newtheorem{property}{Property}
\newtheorem{remark}{Remark}

\newtheoremstyle{style_definition} 
  {} 
  {} 
  {\itshape} 
  {} 
  {\bfseries} 
  {.} 
  { } 
  {\thmname{#1}\thmnumber{ #2}\thmnote{ (#3)}} 

\theoremstyle{style_definition} \newtheorem{definition}{Definition}

\makeatletter
\renewenvironment{proof}[1][\proofname]{\par
\pushQED{\qed}%
\normalfont \topsep6\p@\@plus6\p@\relax
\trivlist
\item\relax
{\itshape
#1\@addpunct{.}}\hspace\labelsep\ignorespaces
}{%
\popQED\endtrivlist\@endpefalse
}
\makeatother

\newif\ifcompile

\newcommand{\myinclude}[1]{\if@preprint
	\include{#1}
\else
  \input{#1}
\fi}

\title{Functional Asplund metrics for pattern matching, robust to variable lighting conditions}
\shorttitle{Functional Asplund metrics}
\shortauthors{Noyel G \etal}

\author[1,2]{Guillaume Noyel}
\author[3,2]{Michel Jourlin}

\email{guillaume.noyel@mines-paris.org,
michel.jourlin@univ-st-etienne.fr}

\affiliation[1]{University of Strathclyde, Department of Mathematics and Statistics, G11XQ Glasgow, Scotland, United Kingdom}

\affiliation[2]{International Prevention Research Institute, 69006 Lyon, France}

\affiliation[3]{Universit\'e Jean Monnet, Laboratoire Hubert Curien, UMR CNRS 5516, 42000 Saint-Etienne, France}

\abstract{In this paper, we propose a complete framework to process images captured under uncontrolled lighting and especially under low lighting.  
By taking advantage of the Logarithmic Image Processing (LIP) context, we study two novel functional metrics: i) the LIP-multiplicative Asplund metric which is robust to object absorption variations and ii) the LIP-additive Asplund metric which is robust to variations of source intensity or camera exposure-time. We introduce robust to noise versions of these metrics. We demonstrate that the maps of their corresponding distances between an image and a reference template are linked to Mathematical Morphology. This facilitates their implementation. We assess  them in various situations with different lightings and movement. Results show that those maps of distances are robust to lighting variations. Importantly, they are efficient to detect patterns in low-contrast images with a template acquired under a different lighting.}


\begin{document}
\begin{paper}


\section{Introduction}
\label{sec:intro}

Metrics or their values, namely the distances, play a central role in image analysis as comparison tools. They possess strong mathematical properties (symmetry, separation, triangular inequality) but they are generally not founded on optical properties. They are therefore not adapted to the comparison of images captured under variable lighting conditions. This issue affects most classical metrics like: the Euclidean-like distances \citep{Li2009,Wang2005}, the integral metric, the uniform metric, the Stepanov distance (or Mink\-owski distance) \citep{Deza2016}, the Hausdorff metric \citep{Dougherty1991,Huttenlocher1993} and many others \citep{Deza2016}. 
The aim of this paper is to propose metric tools robust to lighting variations. 


Analysing images captured with variable lighting conditions is a challenging task that can occur in many settings, such as traffic control \citep{Messelodi2005,Salti2015}, safety and surveillance \citep{Foresti2005}, underwater vision \citep{Peng2017,Ancuti2018}, driving assistance \citep{Hautiere2006}, face recognition \citep{Chen2006,Faraji2016,Shah2015,Lai2014}, large public health databases of images \citep{Noyel2017c}, etc.
There is some data in the literature about the difficulties inherent to this issue.
\citet{Chen2006} observe that the performance of classical techniques like gamma correction \citep{Shan2003}, logarithm transform \citep{Savvides2003}, adaptive histogram equalisation \citep{Pizer1987}, region-based histogram equalisation \citep{Shan2003}, and block-based histogram equalisation \citep{Xie2005} present limitations. They propose a Discrete Cosine Transform in the logarithm domain \citep{Chen2006}. \citet{Faraji2016} suggest other solutions based on logarithmic fractal dimension. \citet{Shah2015} claim that: ``Firstly, textural values are changed during illumination normalisation due to increase in the contrast that changes the original pixels of [images]. Secondly, it minimises the distance between inter-classes which increases the false acceptance rates''. Unsharp masking algorithm \citep{Deng2011} or retinex type of algorithms \citep{Meylan2006} do  not take into account the enhancement of noise. 
Consequently, the usual approach which consists of a pre-processing to normalise the illumination, significantly increases the difficulty to perform the second step dedicated to the recognition of a pattern. Moreover, the pre-processing is rarely based on a rigorous modelling of the cause of the lighting variations. 
We address this issue by proposing metric tools which are efficient in presence of lighting variations, without any pre-processing. Those metrics are especially robust to variations between low-contrast and high-contrast images. We start from a little-known metric defined for binary shapes \citep{Asplund1960,Grunbaum1963}, namely the Asplund metric. It consists of a double-sided probing of one of the shapes by the other. This binary metric has the outstanding property of being \textit{insensitive to object magnification} \citep{Jourlin2014}. Our motivation has been to extend this property to Asplund-like metrics dedicated to grey level images. Such extensions of the binary Asplund metric to the functional case require the use of \textit{homothetic images} or functions which are \textit{insensitive to lighting variations}. Such a concept is mathematically well defined and physically justified in the Logarithmic Image Processing (LIP) framework \citep{Jourlin2001,Jourlin2016}. Two extensions will be studied: i) the LIP-multiplicative Asplund metric based on the LIP-multiplication operation of an image by a real number and ii) the LIP-additive metric based on the LIP-addition operation of an image by a constant. A famous optical law, the Transmittance Law, is at the basis of both LIP-operations \cite[chap. 1]{Jourlin2016}. They give to the functional Asplund metrics a strong physical property: a very low sensitivity to lighting variations, especially for under-lighted images. The LIP-multiplicative Asplund metric is thereby theoretically insensitive to variations of the object absorption (or opacity). The LIP-additive metric is defined to be insensitive to variations of source intensity (or exposure-time). For pattern matching purpose, a map of distances is computed between a template and an image. The closest image patterns to the template correspond to the minimal values of the map. The patterns are then detected by finding the map minima.
In this paper, we will demonstrate that the maps of distances are related to the well-established framework of Mathematical Morphology (MM) \citep{Matheron1967,Serra1982,Heijmans1994,Najman2013}. This will facilitate their programming as numerous image analysis software contain these operations.
Importantly, for the Asplund metrics, there is no empirical pre-processing normalising the image intensity. Moreover, the consistency of the LIP model with Human Vision \citep{Brailean1991} allows the Asplund metrics to perform pattern matching as a human eye would do. The link between these metrics and the LIP model opens the way to numerous applications with low-lighting \citep{Jourlin2016}.

In this paper, our aim is to present a complete framework of pattern matching robust to lighting variations between low-contrast and high-contrast images. In detail, our contribution is two-fold. 
\begin{inparaenum}[(1)]
\item Firstly, we extend the preliminary works defining the functional metrics and their corresponding maps of distances between a template and an image \citep{Jourlin2012,Jourlin2014,Jourlin2016,Noyel2017a,Noyel2017b,Noyel2019}. Beyond the prior work, we add theoretical work at two stages. 
	\begin{inparaenum}[a)]
		\item We better define the robust to noise version of the metrics. 
		\item	We demonstrate the link between the maps of distances and the MM operations of dilation and erosion \citep{Serra1982}. An expression will be given between each map of distances and those MM operations.
	\end{inparaenum}
\item The second part of our contribution is to perform an extensive experimental validation of the metrics on simulated and on real images. We show that the Asplund metrics are efficient for pattern recognition in images captured with different lightings. This is especially the case for the imaging of moving objects, where the motion blur is avoided by shortening the camera exposure-time with the side effect of darkening the images (Fig.~\ref{fig:Moving_disk}).
\end{inparaenum}
\section{Related works and novelty}
\label{sec:rel_wk_novelty}

\subsection{Related Works}
\label{sec:rel_wk_novelty:rel_wk}

Pattern matching methods efficient under variable lighting conditions are seldom studied in the literature. This explains the few references closely related to our work. Nevertheless, we can cite two papers of \citet{Barat2003,Barat2010}. The first one proposes the concept of probing in order to replace one of the following methods. 
\begin{inparaenum}[(i)] 
\item The morphological approach proposed by \citet{Banon1997}. 
\item The extension of the ``hit-or-miss'' transform \citep{Serra1982} to grey-level images introduced by \citet{Khosravi1996}. 
\item The approach inspired by the computation of Hausdorff distance of \citet{Odone2001}. 
\item The ``multi-scale tree matching'' based on boundary segments presented by \citet{Cantoni1998}.
\end{inparaenum} 
In the second paper, \citeauthor{Barat2010} introduce the double-sided probing of a grey level function $f$, where a pair of probes locally surrounds the representative surface of $f$. Such a technique detects the possible locations of the searched pattern. However, the probes are arbitrarily chosen and simply translated along the grey scale to remain in contact with the function $f$. These translations, which darken (or brighten) the probes, seem to take into account the lighting variations of the image. However, as they are not physically founded, they do not correctly model any of such variations.

\subsection{Novelty of the functional Asplund metrics}
\label{sec:rel_wk_novelty:novelty}

The novelty of the functional Asplund metrics is to be theoretically insensitive to lighting variations, with a physical origin. Unlike the related works, the functional Asplund metrics use the LIP-multiplicative and LIP-additive laws to compare a given template to a studied grey level function $f$. Due to the LIP-laws, the template generates itself the pair of probes suited to the values of the function $f$. The arbitrary choice of the probes made by \citet{Barat2010} is therefore avoided in our approach.  


\section{Theoretical background: From a metric for binary shapes to a metric for grey level images}
\label{sec:back}

The Asplund metric originally defined for binary shapes \citep{Asplund1960,Grunbaum1963}, has been extended to grey-level images in the LIP-framework by \citet{Jourlin2012,Jourlin2014}. In this section, we give background notions about the binary Asplund metric, the LIP model, the Functional Asplund metrics and MM.

\subsection{Asplund metric for binary shapes}
\label{sec:back:As_bin}

In the initial definition of the Asplund metric for a pair $(A,B)$ of binary shapes, one shape, e.g. $B$, is chosen to perform the double-sided probing of $A$ by means of two \textit{homothetic} shapes of $B$. Two real numbers are computed: the smallest number $\la_0$ such that $\la_0 B$ contains $A$ and the greatest number $\mu_0$ such that $A$ contains $\mu_0 B$. 
The Asplund distance $d_{asp}(A,B)$ between $A$ and $B$ is then defined according to:
\begin{equation}
	d_{asp} (A,B) = \ln{(\la_0 / \mu_0)}.\label{eq:d_As_bin}%
\end{equation}
\begin{remark} This implies that $d_{asp}$ remains unchanged when one shape is magnified or reduced by any ratio $k$. 
\end{remark}

Fig. \ref{fig_binAs} illustrates the binary Asplund metric where a shape $A$ is probed on its both sides by a reference shape $B$. 

\begin{figure}[!htbp]
\centering
\includegraphics[width=0.49\columnwidth]{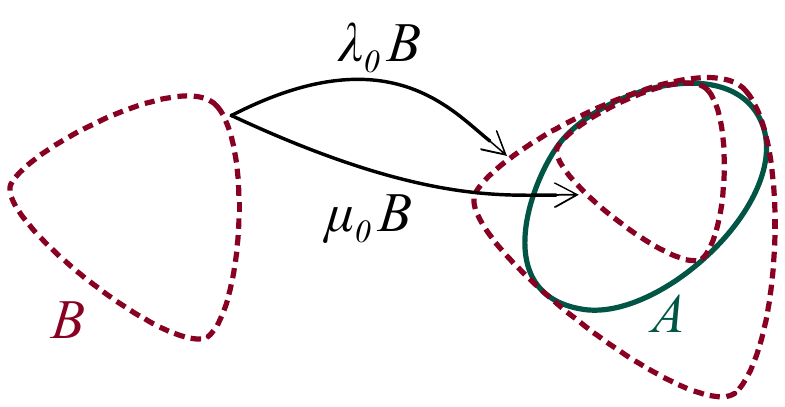}
\caption{Asplund metric for binary shapes. The shape $A$ is probed on both sides by the reference shape $B$ using its two homothetic shapes $\la_0 B$ and $\mu_0 B$.}
\label{fig_binAs}
\end{figure}

The property of insensitivity to object magnification of the binary Asplund metric can be extended to the functional case, where the functional Asplund metric are insensitive to lighting variations. For this purpose, the LIP model is necessary.

\subsection{Logarithmic Image Processing}
\label{sec:back:LIP}

The Logarithmic Image Processing (LIP) model was introduced by \citet{Jourlin1988,Jourlin2001}.
	A grey-level image $f$ is defined on a domain $D \subset \Real^n$ and takes its values in the grey scale $\Tcurv = \left[0,M\right[ \subset \Real$. $f$ is an element of the set $\I = \Tcurv^D$.
	Contrary to common usage, the grey scale extremity $0$ corresponds to the maximal possible intensity observed by the sensor (white), i.e. when no obstacle is located between the source and the sensor. 
	The other extremity of the scale, $M$, corresponds to the limit case where no element of the source is transmitted (black). 
	For 8-bit digitised images, $M$  is equal to 256. Now the two LIP operations on images can be defined: the addition of two images $f \LP g$ and the scalar multiplication $\la \LT f$ of an image by a real number $\la$. Both operations derive from the well-known \textit{Transmittance Law} \citep[chap. 1]{Jourlin2016}: $T_{f \LP g} = T_f \times T_g$.
The transmittance $T_f (x)$ of a grey level image $f$ at a point $x$ of $D$ is the ratio of the out-coming flux at $x$ by the incoming one. The incoming flux corresponds to the source intensity. From a mathematical point of view, the transmittance $T_f (x)$ represents the probability, for a particle of the source arriving in $x$ to be seen by the sensor. The \textit{Transmittance Law} $T_{f \LP g}$ means therefore that the probability of a particle to pass through the addition of two obstacles is nothing but the product of the probabilities to pass through each of them.
The transmittance $T_f (x)$ is related to the grey level $f(x)$ by the equation $T_f(x) = 1 - f(x)/M$.
	By replacing the transmittances $T_f$ and $T_g$ by their expressions in the one of $T_{f \LP g}$, the LIP-addition of two images $f \LP g$ is obtained
\begin{equation}
	f \LP g = f + g - fg/M. \label{eq:LIP:plus}%
\end{equation}
	Considering that the addition $f \LP f$ may be written as $2 \LT f$ according to Eq.~\ref{eq:LIP:plus}, the multiplication of an image $f$ by a real number $\la$ is then expressed as:
\begin{equation}
	\lambda \LT f = M - M \left( 1 - f/M \right)^{\lambda}. \label{eq:LIP:times}%
\end{equation}

\begin{remark} 
The opposite function $\LM f$ of $f$ is easily obtained thanks to the equality $f \LP (\LM f) =0$, as well as the difference between two grey level functions $f$ and $g$:
\begin{align}
	\LM f 	&= (-f)/(1-f/M), \label{eq:LIP:negative}\\
	f \LM g &= (f-g)/(1-g/M). \label{eq:LIP:minus}%
\end{align}
Let us note that $\LM f$ is not an image (as it takes negative values) and $f \LM g$ is an image if and only if $f \geq g$.
\end{remark}

The LIP framework possesses the fundamental properties that are listed hereinafter.
\begin{property}[The LIP framework is not limited to images in transmission]
As the LIP model is consistent with Human Vision \citep{Brailean1991}, the LIP operators are also valid for images acquired in reflected light. 
They simulate the interpretation of images by a human eye.
\end{property}

\begin{property}[Strong physical properties]
For images acquired in transmission, the LIP-addition (or subtraction) of a constant $c$ to (or from) an image $f$ consists of adding (or subtracting) a uniform half-transparent object of grey level $c$, which results in a darkening (or lightening) of $f$. Such operations are useful to correct illuminations variations. The images acquired in transmitted or reflected light have both following properties.
\begin{asparaitem}
	\item The addition (or subtraction) of a constant $c$ to (or from) $f$ simulates the decrease (or increase) of the acquisition exposure-time \citep{Carre2014,Deshayes2015}. If  the values of $f \LM c$ become strictly negative, they behave as light intensifiers \citep[chap. 4]{Jourlin2016}.
	\item The scalar multiplication $\la \LT f$ of $f$ by a positive real number $\la$ signifies that the thickness (or the absorption) of the half-transparent object which generates $f$ is LIP-multiplied by $\la$. The image is darkened if $\la \geq 1$ or lightened if $\la \in \left[0,1\right[$.
\end{asparaitem}
\end{property}

\begin{property}[Strong mathematical properties]
Let $\Fcurv_M=\left]-\infty ,M\right[^D$ be the space of functions defined on $D$ with values in $\left]-~\infty ,M\right[$. $(\Fcurv_M,\LP,\LT)$ is a \textit{real vector space} and the space of images $(\I,\LP,\LT)$ represents its positive cone \citep{Jourlin2001,Jourlin2016}.
\end{property}

In Fig. \ref{fig_LIPstackedSheets}, several half-transparent sheets are stacked upon each other between a light source and a camera \citep{Mayet1996,Jourlin2016}. The camera acquires an image of the light source through the sheets. The perceived intensity $f(x)$ by the camera, at a point $x$, is plotted as a function of the number of sheets. The inverted grey scale is used for the intensity axis. When there is no obstacle, the intensity is 0 (white). With the number of stacked sheets, it increases in a logarithmic way and reaches a maximum $M$ (black), when no light is perceived trough the sheets. The non-linearity of the perceived intensity is taken into account by the LIP model. Such a model will be useful to define functional Asplund metrics with strong properties.

\begin{remark}
There exists a symmetric version of the LIP, namely the Symmetric LIP \citep{Navarro2013}. However, this model is not physically justified albeit it is interesting from a mathematical point of view for its symmetry. It allows e.g. to propose a LIP version of the Laplacian operator or to create Logarithmic Wavelets \citep{Navarro2014}.
\end{remark}

\begin{figure}[!t]
\centering
\includegraphics[width=0.7\columnwidth]{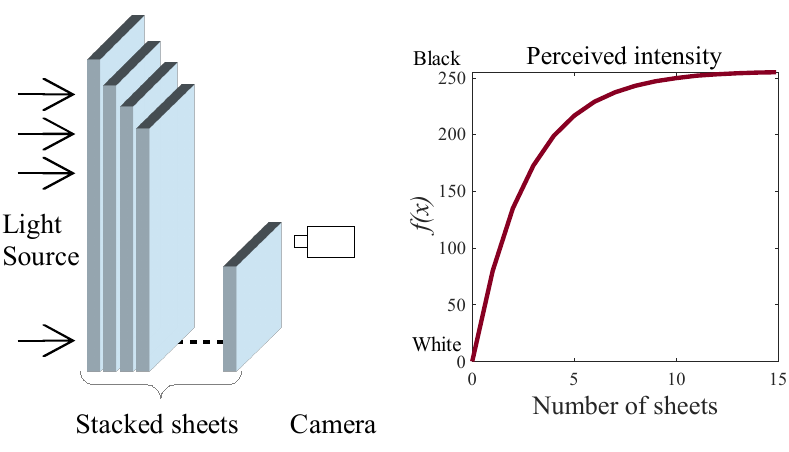}
\caption{A light source passes through several half-transparent sheets which are stacked. The perceived intensity $f(x)$ by the camera is plotted as a function of the sheet numbers using the inverted grey scale.}
\label{fig_LIPstackedSheets}
\end{figure}

\subsection{Definition of Functional Asplund metrics}
\label{sec:back:As_func}

In this section, we will remind the LIP-multiplicative Asplund metric. We will then clarify the definition and properties of the LIP-additive metric.

\subsubsection{LIP-multiplicative Asplund metric}
\label{ssec:back:dAsMult}

Here, we will give the definition of the metric, one of its properties and a rigorous definition. Those were introduced by \citet{Jourlin2012,Jourlin2014,Noyel2017a}. Let $\Tcurv^* = \left]0,M\right[$ be the grey-level axis without the zero value and $\I^*={\Tcurv^*}^D$ the space of images with strictly positive values. 

\begin{definition}[LIP-multiplicative Asplund metric]
Let $f$ and $g \in \I^*$ be two grey level images. As for binary shapes, we select a probing function, e.g. $g$, and we define the two
numbers: $\lambda = \inf \left\{\alpha, f \leq \alpha \LT g \right\}$ and $\mu = \sup \left\{\alpha, \alpha \LT g \leq f\right\}$. 
The LIP-multiplicative Asplund metric $d^{\LT}_{asp}$ is defined by:
\begin{equation}
	d^{\LT}_{asp}(f,g) = \ln \left( \lambda / \mu \right). \label{eq:dasMult}%
\end{equation}
\label{def:dasMult}
\end{definition}

\begin{figure}[!t]
\centering
\includegraphics[width=0.5\columnwidth]{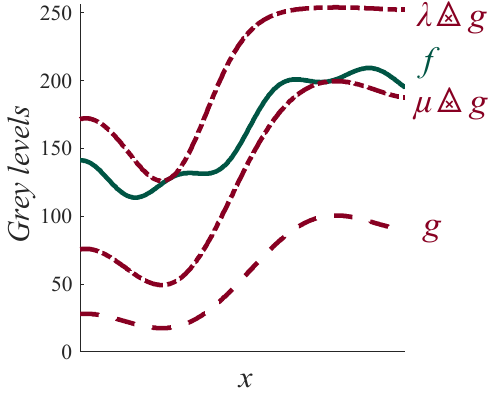}
\caption{Double-sided probing of a function $f$ by a probe $g$ performed by the LIP-multiplicative Asplund metric. $\mu \protect \LT g$ is the lower probe and $\la \protect \LT g$ is the upper probe.}
\label{fig_DoubleProb_DAsMult}
\end{figure}

Fig. \ref{fig_DoubleProb_DAsMult} illustrates the double-sided probing of image $f$ by a probe $g$. The lower probe $\mu \LT g$ is in contact with the lower side of the function $f$ whereas the upper probe $\la \LT g$ is in contact with the upper side of $f$. By comparison with the probe shape, the shapes of the lower and upper probes are deformed by the LIP-multiplication $\LT$ . Such a deformation which depends on the grey value of $g$, is characteristic of the following  metric property.

\begin{property}[Invariance under LIP-multiplication of a constant]
The strongest advantage of the LIP-multiplicative Asplund metric is to remain unchanged when one image (e.g. $f$) is replaced by any homothetic $\alpha \LT f$,
$\forall \alpha \in \Real^{*+}$: $d^{\LT}_{asp}(f,g) = d^{\LT}_{asp}(\alpha \protect \LT f,g)$.
\label{prop:dasMult_invariance}
\end{property}

This latest equation demonstrates the insensitivity of the metric $d^{\LT}_{asp}$ to illumination variations modelled by the LIP-multiplicative law $\LT$, i.e. those which correspond to an absorption change of the object. 

\begin{remark}[\textit{Mathematical appendix \citep[chap. 3]{Jourlin2016}}]
Considering the previous property, it would be more rigorous to explain the LIP-multiplicative Asplund metric as follows.
\begin{asparaenum}[a)]
	\item An equivalence relation $\mathscr{R}$ is defined on the space of grey level images. Two images $f$ and $g\in \I^*$ are said to be in relation if they satisfy: $(f \mathscr{R} g) \Leftrightarrow \exists \alpha > 0, \quad f=\alpha \LT g$.
The previous relation $\mathscr{R}$ obviously satisfies the properties of an equivalence relation (reflexivity, symmetry and transitivity).
	\item To each image $f \in \I^*$, its equivalence class $f^{\LT}$ is associated: $f^{\LT}=\{g,\> g \mathscr{R} f \}$.
	\item A rigorous definition of the LIP-multiplicative Asplund metric is then given into the space of equivalence classes $\I^{\protect \LT}$ by:
	$\forall (f^{\protect \LT},g^{\protect \LT}) \in (\I^{\protect \LT})^2$,\\ $d^{\LT}_{asp}(f^{\protect \LT},g^{\protect \LT}) = d^{\LT}_{asp}(f_1,g_1)$.
	$d^{\LT}_{asp} (f_1,g_1)$ is the Asplund distance between any elements $f_1$ and $g_1$ of the equivalence classes $f^{\LT}$ and $g^{\LT}$.
\end{asparaenum}
\end{remark}

\subsubsection{LIP-additive Asplund metric}
\label{ssec:back:dAsAdd}

As for the LIP-multiplicative metric, we will give a definition of the LIP-additive Asplund metric, some properties and a rigorous definition.

\begin{definition}[LIP-additive Asplund metric]
Let $f$ and $g \in \Fcurv_M $ be two functions, we select a probing function, e.g. $g$, and we define the two
numbers: $c_1 = \inf{ \{c, f \leq c \LP g \}}$ and $c_2 = \sup{ \{c, c \LP g \leq f \} }$, where $c$ lies in the interval $\left]-\infty,M\right[$.
The LIP-additive Asplund metric $d^{\LP}_{asp}$ is defined according to:
\begin{equation}
	d^{\LP}_{asp}(f,g) = c_1 \LM c_2. \label{eq:dasAdd}%
\end{equation}
\label{def:dasAdd}
\end{definition}

\begin{remark}
By definition, the values of $c_1$ and $c_2$ lie in the interval $\left]-\infty,M\right[$, which implies that the probing functions $g \LM c_1$ and $g \LM c_2$ are not always images. However, $c_1$ is always greater than $c_2$. Nevertheless, the Asplund metric has the following property.
\end{remark}

\begin{proposition} 
$d^{\LP}_{asp}(f,g)$ lies in $\left[0,M\right[$ (Proof p. \pageref{proof:prop:dAs_Mult_0_M}).
\label{prop:dAs_Mult_0_M}
\end{proposition}

\begin{property}[Invariance under LIP-addition of a constant]
The LIP-addi\-tive Asplund metric remains unchanged when $f \in \Fcurv_M$ (or $g$) is replaced by any translated function $f \LP k$, with $k \in \left]-\infty,M\right[$. Indeed, the constants $c_1$ and $c_2$ become $c_1 \LP k$ and $c_2\LP k$\linebreak respectively. This implies that $d^{\LP}_{asp}(f \LP k,g) = d^{\LP}_{asp}(f ,g)$ and that $\forall f, \forall k, d^{\LP}_{asp} (f,f \LP k)=0$. Knowing that the addition of a constant to a function is equivalent to a variation of camera exposure-time \citep{Carre2014,Deshayes2015}, we have the fundamental result:
\textit{the LIP-additive Asplund metric is insensitive to exposure-time changing}.
\label{prop:dasAdd_invariance}
\end{property}

\begin{remark}[\textit{Mathematical appendix}]
As for the LIP-mul\-tiplicative Asplund metric a rigorous mathematical definition is obtained by replacing each image $f$ by its equivalence class $f^{\LP}$, which represents the set of functions $h$ such that $h=f \LP k$, for a constant $k$ lying in $\left]-\infty,M\right[$. Nevertheless, if we keep the notation $f$ and $g$, there is no ambiguity. Indeed, for a couple $(f^{\protect \LP},g^{\protect \LP})$ of equivalence classes, we have $d^{\LP}_{asp}(f^{\protect \LP},g^{\protect \LP}) = d^{\LP}_{asp}(f_1,g_1)$,
	where $f_1$ and $g_1$ are elements of the classes $f^{\LP}$ and $g^{\LP}$ respectively.
\end{remark}


\subsection{Fundamental operations in Mathematical Morphology}
\label{sec:back:MM}

As the definitions of the functional Asplund metrics are based on extrema between functions, there exists a natural link with MM as shown by \citet{Noyel2017a}. MM \citep{Matheron1967,Serra1982,Heijmans1990,Soille2003,Schonfeld2008,Najman2013,vanDeGronde2014} is defined in complete lattices \citep{Serra1982,Banon1993}. Let us recall some important definitions.

\begin{definition}[Complete lattice]
Given a set $\Lscr$ and a partial order $\leq$ on $\Lscr$, $\Lscr$ is a complete lattice if every subset $\mathscr{X}$ of $\Lscr$ has an infimum (a greatest lower bound) and a supremum (a least upper bound).
\end{definition}
The infimum and the supremum of $\mathscr{X}$ will be denoted by $\bigwedge \mathscr{X}$ and $\bigvee \mathscr{X}$, respectively. Two elements of the complete lattice $\Lscr$ are important: the least element $O$ and the greatest element $I$. E.g. the set of images from $D$ to $[0,M]$, $\overline{\I}=[0,M]^D$, is a complete lattice with the partial order relation $\leq$. 
The least and greatest elements are the constant functions $f_0$ and $f_M$ whose values are equal to $0$ and $M$, respectively, for all the elements of $D$. The supremum and infimum are defined by taking the pointwise infimum and supremum, respectively. For $\mathscr{X} \subset \overline{\I}$, we have $(\bigwedge_{\overline{\I}} \mathscr{X})(x) = \bigwedge_{[0,M]} \{ f(x):f \in \mathscr{X}, \> x\in D \}$ and $(\bigvee_{\overline{\I}} \mathscr{X})(x)  = \bigvee_{[0,M]} \{ f(x):f \in \mathscr{X}, \> x\in D \}$.
Given $\overline{\Real} = \Real \cup \{-\infty , +\infty \}$, the set of functions $\overline{\Real}^D$ is also a complete lattice with the usual order $\leq$.

\begin{definition}[Erosion, dilation \citep{Banon1993}]
Given $\Lscr_1$ and $\Lscr_2$ two complete lattices, a mapping $\psi \in \Lscr_2^{\Lscr_1}$ is
\begin{itemize}
	\item an erosion $\varepsilon$: iff $\forall \mathscr{X} \subset \Lscr_1$, $\psi( \bigwedge \mathscr{X} ) =  \bigwedge \psi( \mathscr{X} )$;
	\item a dilation $\delta$:\hphantom{:} iff $\forall \mathscr{X} \subset \Lscr_1$, $\psi( \bigvee \mathscr{X} ) =  \bigvee \psi( \mathscr{X} )$.
\end{itemize}
As the definitions of these mappings apply even to the empty subset of $\Lscr_1$, we have: $\varepsilon(I)=I$ and $\delta(O)=O$.
\label{pre:def_morpho_base}
\end{definition}

A structuring function $b$ is a function defined on the domain $D_b \subset D$ with its values in $\Tcurvb = \Realb$, or in $\Tcurvb = [0,M]$. In the case of a dilation or an erosion of a function $f$ by an additive structuring function $b$, which is invariant under translation in the domain $D$, the previously defined dilation $\delta$ or erosion $\varepsilon$ can be expressed in the same lattice $(\Realb^D,\leq)$, or $(\overline{\I},\leq)$ \citep{Serra1988,Heijmans1990} by:
%
\begin{align}
(\delta_b(f))(x)			&= \bigvee_{h \in D_b} 	 \left\{ f(x - h) + b(h)\right\}\nonumber\\  
											&= (f \oplus b) (x), \label{eq:dilate_funct}\\
(\varepsilon_b(f))(x) &= \bigwedge_{h \in D_b}  \left\{ f(x + h) - b(h) \right\}\nonumber\\ 
											&= (f \ominus b) (x). \label{eq:erode_funct}%
\end{align}
The symbols $\oplus$ and $\ominus$ represent the extension to functions \citep{Serra1982} of Minkowski operations between sets \citep{Serra1982}. The term ``additive structuring function'' refers to a vertical translation in the image space $\Tcurvb$ \citep{Heijmans1990}. 



\section{Pattern analysis with the LIP-multiplicative Asplund metric}
\label{sec:AsLIPmult}

The LIP-multiplicative Asplund metric is of utmost importance for pattern matching of objects whose absorption (or opacity) is varying.
For this purpose, \citet{Jourlin2012} introduced a map of LIP-multiplicative Asplund distances. \citet{Noyel2015} studied it and \citet{Noyel2017a,Noyel2017b}\linebreak established its link with MM through different conference papers. In this section, we will recall its definition, we will improve the definition of its robust to noise version and we will clearly present the link with MM.

\begin{figure}[!t]
\centering
\subfloat[]{\includegraphics[width=0.40\columnwidth]{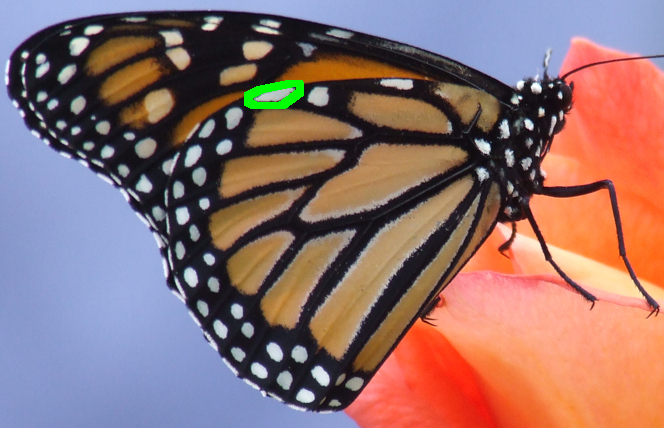}%
\label{fig_MapAsplundMult_Butterfly:1}}
\hfil
\subfloat[]{\includegraphics[width=0.40\columnwidth]{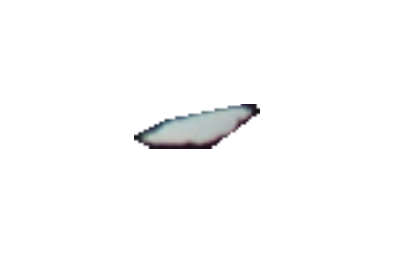}%
\label{fig_MapAsplundMult_Butterfly:2}}
\hfil
\subfloat[]{\includegraphics[width=0.40\columnwidth]{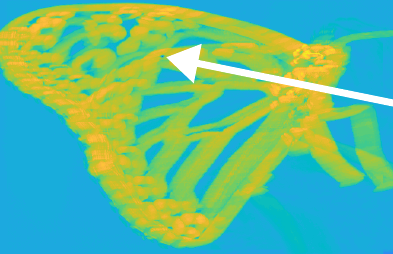}%
\label{fig_MapAsplundMult_Butterfly:3}}
\hfil
\subfloat[]{\includegraphics[width=0.40\columnwidth]{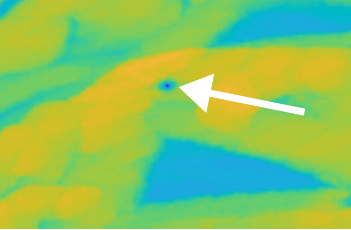}%
\label{fig_MapAsplundMult_Butterfly:4}}
\caption{(a) Colour version of the butterfly image. In its luminance image $f$, a white spot -- surrounded by the green curve -- is selected as a probe function $b$. (b) Zoom in on the colour version of the luminance probe $b$. (c) Map of LIP-multiplicative Asplund distances $Asp_{b}^{\protect \LT} f$ between the image $f$ and the probe $b$. Its minimum is indicated by the white arrow. (d) Zoom in on the map of Asplund distances of the image. The white arrow points to its minimum.}
\label{fig_MapAsplundMult_Butterfly}
\end{figure}

\subsection{Map of Asplund distances}
\label{sec:AsLIPmult:map}

Let $\Tcurv = \left[0,M\right[$, $\Tcurv^* = \left]0,M\right[$ be grey-level axes and $\I^*={\Tcurv^*}^D$, $D \in \Real^n$ the space of strictly positive images.

\begin{definition}[Map of LIP-multiplicative Asplund distances \citep{Jourlin2012}]
Let  $f \in \I^*$  be a grey-level image and $b \in (\Tcurv^*)^{D_{b}}$ a probe. The map of Asplund distances $Asp_{b}^{\LT}: \I^* \rightarrow \left(\Real^{+} \right)^{D}$ is defined by:
\begin{align}
Asp_{b}^{\LT} f(x) &= d^{\LT}_{asp} (f_{\left|D_b(x)\right.},b).\label{eq:map_AsMult}%
\end{align}
\label{def:map_As_mult}
\end{definition}
For each point $x \in D$, the distance $d^{\LT}_{asp} (f_{\left|D_b(x)\right.},b)$ is computed in the neighbourhood $D_b(x)$ centred in $x$ and the template $b$ is acting like a structuring element. $f_{\left|D_b(x)\right.}$ is the restriction of $f$ to $D_b(x)$.  $Asp_{b}^{\LT} f: D \rightarrow \Real^{+}$ is the map of Asplund distances between the image $f$ and the probe $b$.


Fig. \ref{fig_MapAsplundMult_Butterfly} illustrates the map of Asplund distances with an image of a butterfly \citep{YFCC100M_butterfly2010}. The image is coming from the Yahoo Flickr Creative Commons 100 Million Dataset \citep{Thomee2016}. The processing is performed on its luminance image $f$ which is in grey levels even if the images are displayed in colours. In the luminance image $f$ of the butterfly (Fig. \ref{fig_MapAsplundMult_Butterfly:1}), a white spot is selected to serve as a probe $b$ (Fig. \ref{fig_MapAsplundMult_Butterfly:2}). The map of Asplund distances $Asp_{b}^{\LT} f$ between the image $f$ and the probe $b$ (Fig. \ref{fig_MapAsplundMult_Butterfly:3}) presents a minimum which corresponds to the probe we are looking for (Fig. \ref{fig_MapAsplundMult_Butterfly:4}). A map of Asplund distances allows us therefore to find a reference pattern or probe within an image. However, as images may present acquisition noise, a robust to noise version of the metric is useful for pattern matching.

\subsection{A robust to noise version}
\label{ssec:AsLIPmult:rob_noise}

The Asplund metric is computed using extrema, which makes it sensitive to noise. To overcome this limitation, \citet{Jourlin2014} have proposed an extension which removes  from $D$ the most penalising points. This idea is related to the \textit{topology of the measure convergence} \cite[chap. 4]{Bourbaki2007_integration_chap_1_4} in the context of grey-level images.
As the image is digitised, the number of pixels lying in $D$ is finite. The ``measure'' of a subset of $D$ is linked to the cardinal of this subset, e.g. the percentage $P$ of its elements related to the domain $D$ (or a region of interest $R\subset D$). 

%

We are looking for a subset $D'$ of $D$, such that $i)$ $f_{\left|D'\right.}$ and $g_{\left|D'\right.}$ are neighbours (for the Asplund metric) and $ii)$ the complementary set $D\setminus D'$ of $D'$ related to $D$ is small-sized as compared to $D$. This last condition is written: $P(D \setminus D') = \frac{\#(D \setminus D')}{\#D} \leq p$, where $p$ represents an acceptable percentage and $\#D$ the number of elements in $D$. A neighbourhood of the image $f \in \I$ can be defined thanks to a small positive real number $\epsilon$ as: 
	\begin{equation}
	\setlength{\nulldelimiterspace}{0pt}
	\begin{array}{@{}r@{\hspace{0.2em}}c@{\hspace{0.2em}}l@{}}
		N_{P,d^{\LT}_{asp},\epsilon,p}(f) &=& \left\{ g \setminus \exists D' \subset D,\> d^{\LT}_{asp}(f_{\left|D'\right.},g_{\left|D'\right.}) < 			\epsilon \right.\\
		&&\left. \text{ and } P(D \setminus D') \leq p\right\}.%
		\end{array}
		\label{eq:neighbourhood_tolerance_As_mult}
	\end{equation}

The set $D'$ corresponds to the noise pixels to be discarded. As these pixels are the closest to the probe, they are selected by a function $\gamma^{\LT}_{(f,g)}:~D~\rightarrow~\Real$ characterising the (LIP-multiplicative) contrast between the functions $f$ and $g$ and which is defined by:
\begin{align}
	\forall x \in D,\> \gamma^{\LT}_{(f,g)}(x) \LT g(x) &= f(x) \nonumber\\
	\Leftrightarrow \gamma^{\LT}_{(f,g)} &= \frac{\ln{\left(1-f/M\right)}}{\ln{\left(1-g/M\right)}}. \label{eq:LM_contrast_fct}%
\end{align}
$\gamma^{\LT}_{(f,g)}(x)$ is the real value by which each image value $f(x)$ is LIP-multiplied to be equal to the probe value $g(x)$. 
The expressions of the probes $\la \LT g$ and $\mu \LT g$ (Def. \ref{def:dasMult}) can be written with the contrast function $\gamma^{\LT}_{(f,g)}$ as follows: $\la = \inf{\{\alpha, \forall x, \gamma^{\LT}_{(f,g)}(x) \leq \alpha \}}$ and\linebreak $\mu = \sup{\{\alpha, \forall x, \alpha \leq}$ ${\gamma^{\LT}_{(f,g)(x)}\}}$. This explains that the closest image values to the upper probe $\la \LT g$, or the lower probe $\mu \LT g$, correspond to the greatest, or smallest, values of $\gamma^{\LT}_{(f,g)}$, respectively. Using this property, new probes $\la' \LT g$ , or $\mu' \LT g$, can be defined on the domain $D\setminus D'$ obtained by discarding a percentage $(1-p)/2$ of the pixels with the greatest, or smallest, contrast values $\gamma^{\LT}_{(f,g)}(x)$, respectively. The restricted domain $D\setminus D'$  has thereby a cardinal equal to a percentage $p$ of the cardinal of $D$.

\begin{definition}[LIP-multiplicative Asplund metric with tolerance]
Let $(1-p)$ be a percentage of points of $D$ to be discarded and $D'$ the set of these discarded points. The (LIP-multiplicative) Asplund metric with tolerance between two grey-level images $f$ and $g \in \I$ is defined by:
\begin{align}
d^{\LT}_{asp,p} (f,g) &= \ln( \la' / \mu' ). \label{eq:dAsMult_tol}%
\end{align}
The factors $\la'$ and $\mu'$ are equal to\linebreak $\la' = \inf \{\alpha, \forall x \in D,$ $\gamma^{\LT}_{(f_{ |D \setminus D'},g_{|D \setminus D'})}(x) \leq \alpha \}$ and\linebreak $\mu'  = \sup \{ \alpha , \forall x \in D, \alpha \leq \gamma^{\LT}_{(f_{|D \setminus D'},g_{|D \setminus D'})}(x) \}$. A percentage $(1-p)/2$ of the points $x\in D$ with the greatest, respectively lowest, contrast values $\gamma^{\LT}_{(f,g)}(x) = \frac{\ln{(1-f(x)/M)}}{\ln(1-g(x)/M)}$ are discarded.
\label{def:dAsMult_tol}
\end{definition}

Using Eq.~\ref{eq:LM_contrast_fct} and \ref{eq:LIP:times}, the contrast function $\gamma^{\LT}_{(f,g)}$ is proportional to the contrast function $\gamma^{\LT}_{(f,\alpha \LT g)}$ between the image $f$ and the LIP-multiplied probe $\alpha \LT g$: $\gamma^{\LT}_{(f,\alpha \LT g)}(x) = (1/\alpha) \gamma^{\LT}_{(f,g)}(x)$, with $\alpha >0$. This leads to the following property which is demonstrated in the supplementary materials.

\begin{property}
\label{prop:inv_LIPMult_dAsMult_tol}
The metric $d^{\LT}_{asp,p}$ is invariant under LIP-multi\-plication by a scalar.
\end{property}

\begin{remark}
A map of Asplund distances with tolerance can be defined as in definition \ref{def:map_As_mult}: $Asp_{b,p}^{\LT}f(x) = d^{\LT}_{asp,p} (f_{|D_b(x)},b)$.
\end{remark}

Fig. \ref{fig_AsplundMult:1} illustrates the ability of the Asplund metric with tolerance to discard the extremal values $f(x_{\la})$ and $f(x_{\mu})$ associated to noise. In Fig. \ref{fig_AsplundMult:2}, the probe $g$ is chosen as a plane and the image $f$ is obtained by adding a noise to the probe with a spatial density of $\rho = 8\%$. The plane of the probe $g$ is distorted by the LIP-multiplication used for the upper probe $\la \LT g$ and for the lower probe $\mu \LT g$. The Asplund distance which is initially equal to $d^{\LT}_{asp}(f,g) = 3.75$ decreases to $d^{\LT}_{asp,p}(f,g) = 0.79$, with a tolerance of $p=97\%$. Experiments have shown that with $p=90 \%$ ($\leq (1-\rho)$), the Asplund distance is equal to zero.\pagebreak 

\begin{figure}[!t]
\centering
\subfloat[]{\includegraphics[width=0.49\columnwidth]{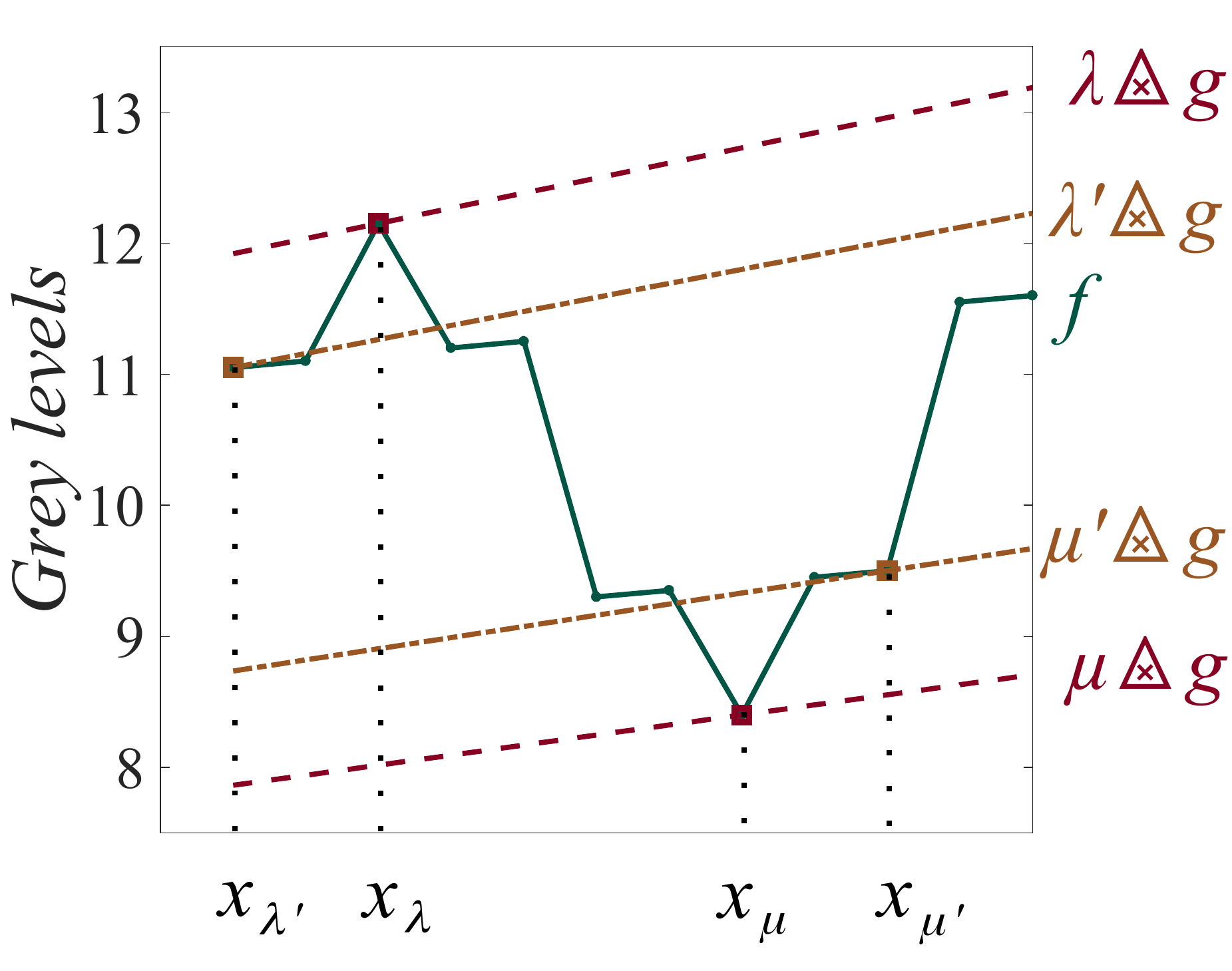}%
\label{fig_AsplundMult:1}}
\hfil
\subfloat[]{\includegraphics[width=0.49\columnwidth]{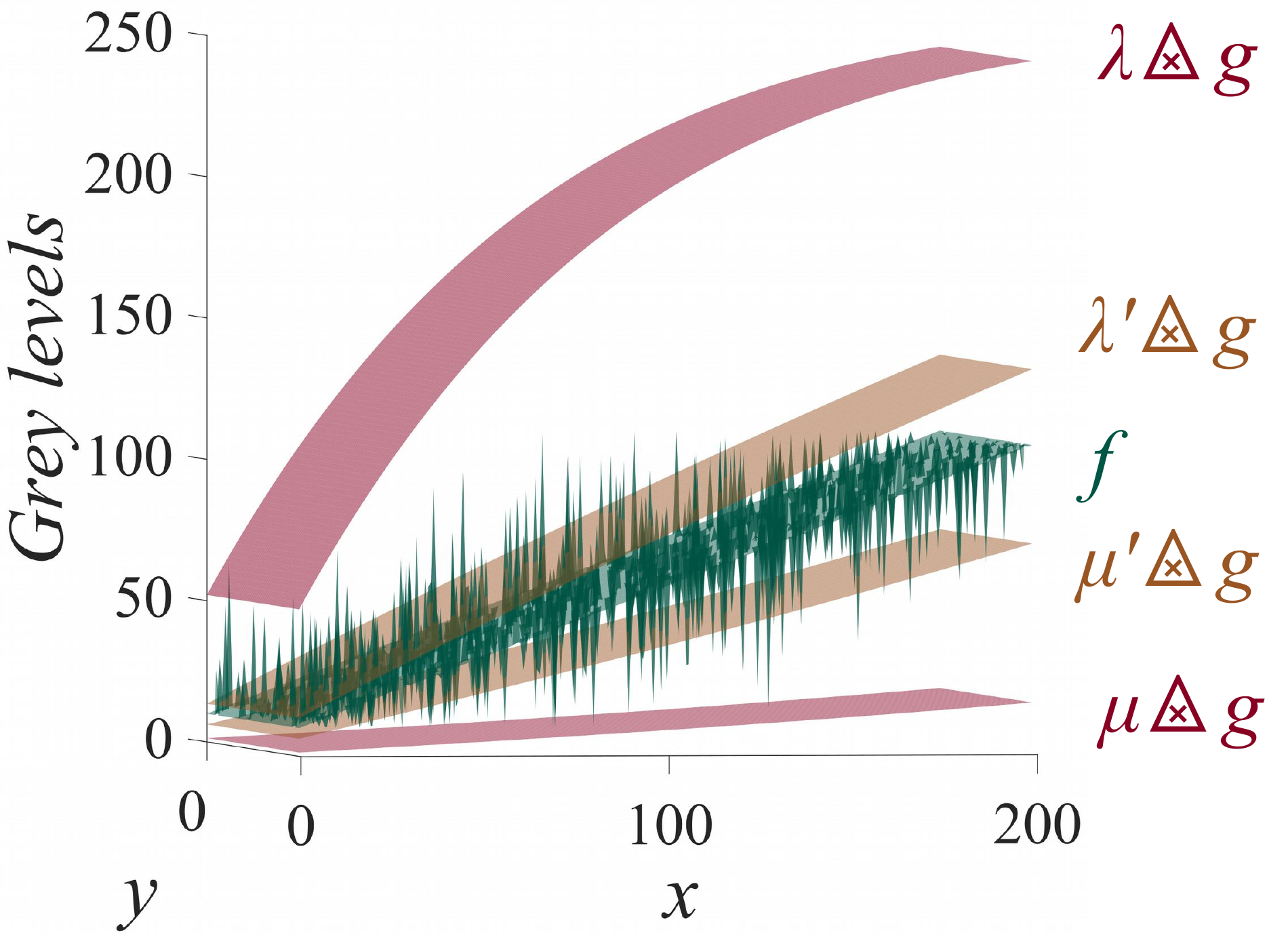}%
\label{fig_AsplundMult:2}}
\caption{Illustration of the LIP-multiplicative Asplund metric where $g$ is used to probe $f$. $\lambda {\protect \LT} g$ (respectively $\la' {\protect \LT} g$) is the upper probe and $\mu {\protect \LT} g$ (resp. $\mu' {\protect \LT} g$) is the lower probe of the Asplund metric $d^{\protect \LT}_{asp}(f,g)$ (resp. of the Asplund metric with a tolerance $p$, $d^{\protect \LT}_{asp,p}(f,g)$). (a) Without tolerance, the Asplund distance is equal to $d^{\protect \LT}_{asp}(f,g) = 0.42$ whereas with a tolerance $p=80\%$ it decreases to $d^{\protect \LT}_{asp,p}(f,g) = 0.24$. (b) The image $f$ is obtained by adding to the probe $g$ a Gaussian white noise of mean 0, variance 5 and spatial density $\rho~=~8\%$.}
\label{fig_AsplundMult}
\end{figure}

\subsection{Link with Mathematical Morpho-logy}
\label{ssec:AsLIPmult:MM}
We first present the maps of distances with neighbourhood operations. We then establish the link with MM in the specific case of a flat structuring element and in the general case of a structuring function.


\subsubsection{General expression of the map of Asplund distances with neighbourhood operations}
\label{sssec:AsLIPmult:MM:gen_expr}


From Eq.~\ref{eq:dasMult}, for each $x\in D$, there is $d^{\LT}_{asp} (f_{\left|D_b(x)\right.},b) = \ln{ \left( \la_b f(x) / \mu_b f(x) \right)}$ with \linebreak $\lambda_b f(x) = \inf \left\{\alpha, f_{\left|D_b(x)\right.} \leq \alpha \LT b \right\}$ and $\mu_b f(x) = \sup \left\{\alpha,\right.$ $\left.\alpha \LT b \leq f_{\left|D_b(x)\right.}\right\}$. These expressions show that the map of Asplund distances consists of a double-sided probing, at each point $x$, by the least upper bound $\lambda(x) \LT b$ and by the greatest lower bound $\mu(x) \LT b$. As $\la_b f(x)$ and $\mu_b f(x)$ exist for all $x \in D$, the following maps can be defined \citep{Noyel2017a}.

\begin{definition}[LIP-multiplicative maps of the least upper and of the greatest lower bounds \citep{Noyel2017a}]
Given $\overline{\Real}^+=[0,+\infty]$, let $f \in \overline{\I}$ be an image and $b \in (\Tcurv^*)^{D_{b}}$ a probe. Their map of the least upper bounds (mlub) $\la_b: \overline{\I} \rightarrow (\overline{\Real}^{+} )^{D}$ and their map of the greatest lower bounds (mglb) $\mu_{b}: \overline{\I} \rightarrow (\overline{\Real}^{+} )^{D}$ are defined by:
\begin{align}
		\la_{b} f(x) &=  \inf_{h \in D_b}{ \{ \alpha, f(x+h) \leq \alpha \LT b(h) \}}, \label{eq:upper_map}\\
		\mu_{b} f(x) &=  \sup_{h \in D_b}{ \{ \alpha, \alpha \LT b(h) \leq f(x+h) \}}. \label{eq:lower_map}%
	\end{align}
\end{definition}

Let us define $\tilde{f}= \ln{\left( 1 - f/M \right)}$, $f \in \Ib$ and introduce the general expression of the mlub and of the mglb.
Propositions \ref{prop:gen_expr_lower_upper_maps} to \ref{prop:property_lower_upper_maps} are demonstrated in \citep{Noyel2017a} and in the supplementary materials. Proposition \ref{prop:expr_mapAsMult_morpho} is demonstrated in the \nameref{sec:app}.

\begin{proposition}
	The mlub $\la_{b}$ and mglb $\mu_{b}$ are equal to:
	\begin{align}
	\la_{b} f(x) &= \bigvee{ \{ \tilde{f}(x+h) / \tilde{b}(h) , h \in D_b \} },   \label{eq:upper_map_2}\\
	\mu_{b} f(x) &= \bigwedge{ \{ \tilde{f}(x+h) / \tilde{b}(h) , h \in D_b \} }. \label{eq:lower_map_2}%
	\end{align}
	\label{prop:gen_expr_lower_upper_maps}
\end{proposition}

\begin{corollary}
	Given $f \in \overline{\I}$, $f > 0$, the map of LIP-multiplicative Asplund distances becomes:
	\begin{align}
	Asp_{b}^{\LT}f &= \ln \left( \frac{ \la_{b} f }{ \mu_{b} f }\right).\label{eq:map_AsMult_la_mu_general_se}%
	\end{align}
	\label{corol:gen_expr_map_As}
\end{corollary}

\subsubsection{Particular case of a flat structuring element}
\label{sssec:AsLIPmult:MM:flat_se}

In the case of a flat structuring element, the expressions of the different maps can be simplified as follows. 
\begin{proposition}
	Let $b=b_0 \in (\Tcurv^*)^{D_b}$ be a flat structuring element ($\forall x \in D_b$, $b(x) = b_0$). The mlub $\la_{b_0}$, the mglb $\mu_{b_0}$ and the map of Asplund distances $Asp_{b_0}^{\LT}$ are equal to:
	\begin{align}
		\la_{b_0} f &= (1/\tilde{b}_0) \ln{ [ 1 - ( \delta_{\overline{D}_b} f ) / M ] }, \label{eq:upper_map_flat_se}\\
		\mu_{b_0} f &= (1/\tilde{b}_0) \ln{ [ 1 - ( \varepsilon_{D_b} f ) / M ] }, \label{eq:lower_map_flat_se}\\
		Asp_{b_0}^{\LT} f &= \ln{ \left[ \frac{ \ln{ [ 1 - (\delta_{\overline{D}_b} f) / M ] } }{ \ln{ [ 1 - (\varepsilon_{D_b} f) / M ] } }\right] }, \> \text{ where } f > 0.\label{eq:map_AsMult_la_mu_flat_se}%
	\end{align}
	$\overline{D}_b = \{-h, h\in D_b \}$ is the reflected (or transposed) domain of the structuring element.
	\label{prop:expr_maps_flat_se}
\end{proposition}

This result shows that the map of Asplund distances is a combination of logarithms, an erosion $\varepsilon_b$ and a dilation $\delta_b$ of the image $f$ by the flat structuring element $b$. As numerous image processing libraries include erosion and dilation operations, the program implementation becomes easier.

\subsubsection{General case: a structuring function}
\label{sssec:AsLIPmult:MM:struc_fct}


\begin{proposition}
	The mlub $\la_b$ and the mglb $\mu_b$ are a dilation and an erosion, respectively, between the two complete lattices $\Lscr_1 = (\overline{\I} , \leq)$ and $\Lscr_2 = ((\overline{\Real}^{+})^{D},\leq)$.
	\label{prop:property_lower_upper_maps}
\end{proposition}
Let us determine the expressions of this dilation and this erosion using MM with multiplicative structuring functions introduced by \citet{Heijmans1990,Heijmans1994}.

\begin{definition}[Erosion and dilation with a multiplicative structuring function \citep{Heijmans1990,Heijmans1994}]
Given a function $f \in (\overline{\Real}^{+})^{D}$, and $b \in (\overline{\Real}^{+})^{D_b}$ a multiplicative structuring function:
\begin{align}
\bigvee_{h \in D_b} 	 \left\{ f(x - h) . b(h) \right\} &= (f \dot{\oplus} b) (x)  \label{eq:dilation_funct_mult}
\end{align}
is a dilation and
\begin{align}
\bigwedge_{h \in D_b} \left\{ f(x + h) / b(h) \right\} &= (f \dot{\ominus} b) (x)  \label{eq:erosion_funct_mult}%
\end{align}
is an erosion, with the convention that $f(x-h).b(h)=0$ when $f(x-h)=0$ or $b(h) =0$ and that $f(x+h)/b(h)= + \infty$ when $f(x+h) = + \infty$ or $b(h) = 0$. The symbols $\dot{\oplus}$ and $\dot{\ominus}$ represent the extension to multiplicative structuring functions of Minkowski operations between sets \citep{Serra1982}.
\end{definition}

There exists a relation between the multiplicative erosion or dilation and the additive operations of section \ref{sec:back:MM} \citep{Heijmans1990,Heijmans1994}:
\begin{align}
	f \dot{\oplus} b  &=  \exp{ (\ln f \oplus \ln b )}, \label{eq:rel_dil_mult_add}\\
	f \dot{\ominus} b &=  \exp{ (\ln f \ominus \ln b )}. \label{eq:rel_ero_mult_add}
\end{align}

The next proposition gives the morphological expressions of the different maps.
\begin{proposition}
	Let $b\in \Tcurv^{D_b}$ and $f \in \overline{\I}$, the expressions of the mlub $\la_{b}$, which is a dilation, and of the mglb $\mu_{b}$, which is an erosion, are:
	\begin{alignat}{3}
		\la_{b} f &= (-\tilde{f}) \dot{\oplus} (-1/\tilde{\overline{b}}) 	&&= \exp{ ( \hat{f} \oplus (- \hat{\overline{b}}) ) }, \label{eq:upper_map_morpho}\\
		\mu_{b} f &= (-\tilde{f}) \dot{\ominus} (-\tilde{b}) 							&&= \exp{ ( \hat{f} \ominus \hat{b} ) }. \label{eq:lower_map_morpho}%
	\end{alignat}
$\overline{b}$ is the reflected structuring function defined by $\forall x \in \overline{D}_b$, $\overline{b}(x)=b(-x)$ \citep{Soille2003} and $\tilde{f}$ is the function $\tilde{f} = \ln{\left( 1 - f/M \right)}$, with $\tilde{f} \in \left[-\infty,0\right]$. $\hat{f} = \ln{(- \tilde{f})} = \ln{(- \ln{\left( 1 - f/M \right)})}$ is a transform of $f$.
	
The map of Asplund distances is the difference between a dilation and an erosion whose expression is:
	\begin{align}
		Asp_{b}^{\LT}f &= \left[ \hat{f} \oplus (- \hat{\overline{b}}) \right] - \left[ \hat{f} \ominus \hat{b} \right]
	= \delta_{-\hat{\overline{b}}} \hat{f} - \varepsilon_{\hat{b}} \hat{f}.\label{eq:As_mapMult_morpho}%
	\end{align}
	\label{prop:expr_mapAsMult_morpho}
\end{proposition}
We notice that the map of LIP-multiplicative Asplund distances is similar to the norm of the morphological gradient $\varrho_{\hat{b}}$ also named Beucher's gradient \citep{Serra1988,Soille2003}. It is the difference between a dilation and an erosion of the transformed image $\hat{f}$ by a structuring function $\hat{b}$: $\varrho_{\hat{b}} \hat{f} = \delta_{\hat{b}} \hat{f} - \varepsilon_{\hat{b}} \hat{f}$.
This similarity shows that the map of Asplund distances acts as an operator of derivation.

\begin{remark} The map of Asplund distances with tolerance $Asp_{b,p}^{\LT}$ can be computed by replacing the dilation and erosion by rank-filters \citep{Serra1988}. 
\end{remark}

Fig. \ref{fig_MapAsplundMult_signal:1} illustrates the double-sided probing of an image $f$ by a probe $b$ using the LIP-multiplicative law $\LT$ which modifies the amplitude of the upper probes $\la_b f(x) \LT b$ and of the lower probes $\mu_b f(x) \LT b$. Both peaks have a different amplitude caused by a lighting drift created with the LIP-multiplicative law. In Fig. \ref{fig_MapAsplundMult_signal:2}, when the probe $b$ is similar to a pattern in $f$ (according to the Asplund distance), the map of Asplund distances of $f$, $Asp_b^{\LT} f$, presents a local minimum. Here, both peaks are located at the deepest minima of the map of $f$ by the probe $b$, $Asp_b^{\LT} f$. This result shows that the map of LIP-multiplicative Asplund distances is insensitive to a lighting drift corresponding to a variation of absorption (or opacity) of the object.

\begin{figure}[!t]
\centering
\subfloat[]{\includegraphics[width=0.49\columnwidth]{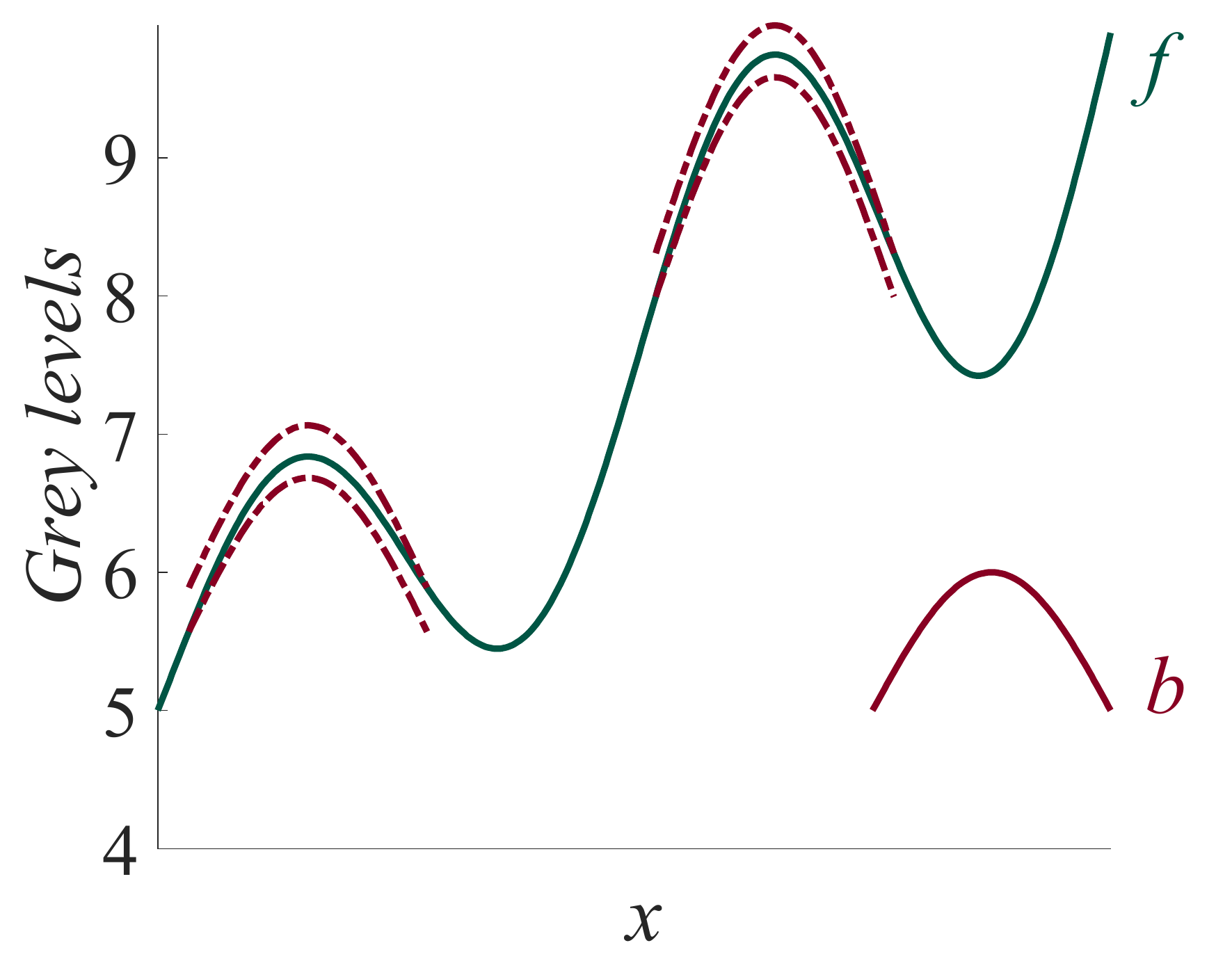}%
\label{fig_MapAsplundMult_signal:1}}
\hfil
\subfloat[]{\includegraphics[width=0.49\columnwidth]{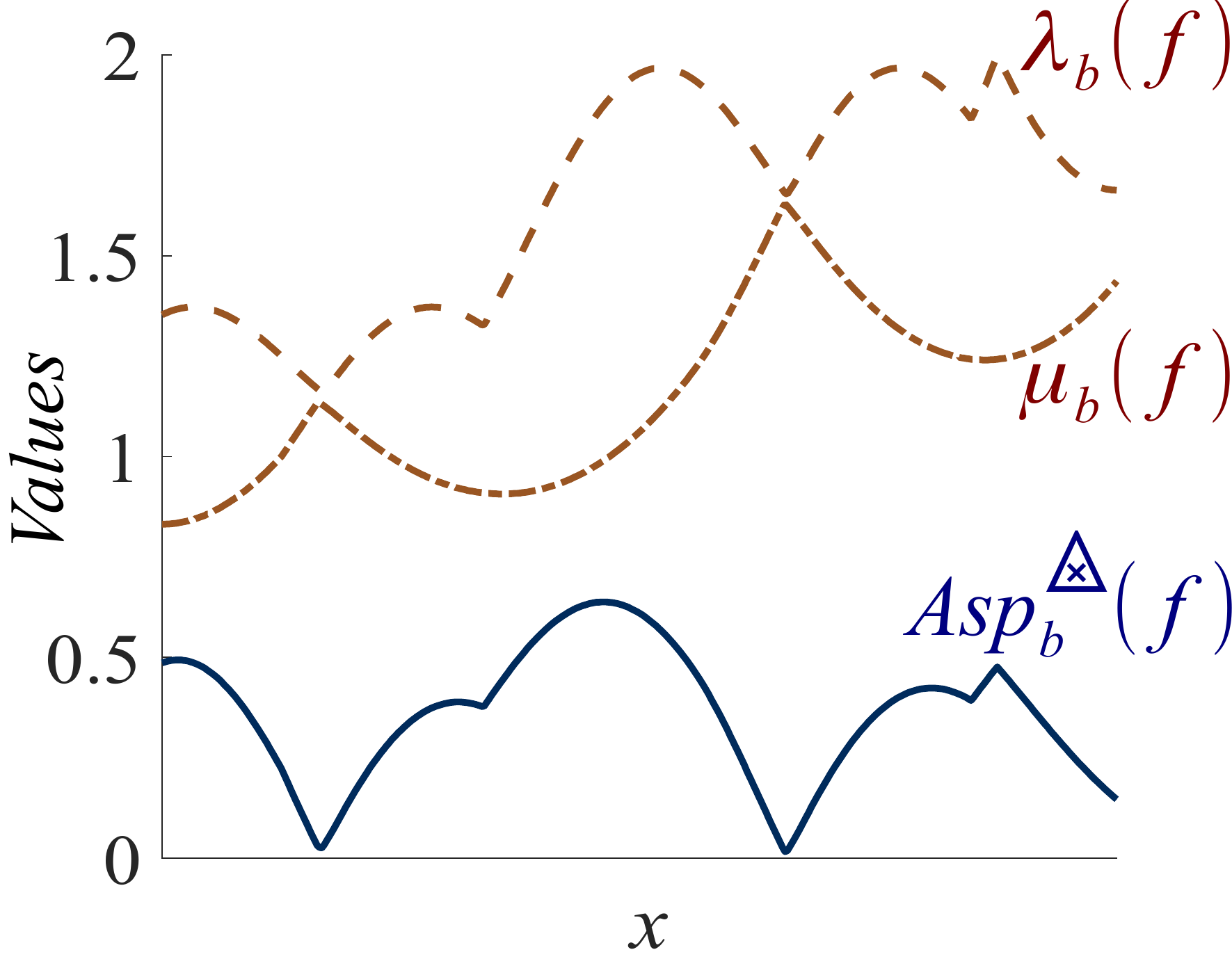}%
\label{fig_MapAsplundMult_signal:2}}
\caption{(a) The image $f$ is probed on both sides by the probe $b$ (in red colour) using the LIP-multiplicative law ${\protect \LT}$. The double-probing is depicted at two locations by the red dashed curves. (b) $\la_b f$ and $\mu_b f$ are the mlub and mglb of $f$, respectively. $Asp_b^{{\protect \LT}} f$ is the map of Asplund distances between $f$ and $b$.}
\label{fig_MapAsplundMult_signal}
\end{figure}


\section{Pattern analysis with the LIP-additive Asplund-like  metric}
\label{sec:AsLIPadd}

The LIP-additive Asplund metric is useful for images acquired with a small source intensity or a short exposure-time \citep[chap. 3]{Jourlin2016}.
In this section, we will introduce: a map of LIP-additive Asplund distances, a robust to noise version of the metric and the link between the map of distances and MM. Due to the important similarity with the LIP-multiplicative case (see previous section), we will only point out the new equations and results. 
We will consider the set of functions $\Fcurv_M=\Tcurv^D$ (or $\Fcurvb_M=\Tcurvb^D$) with values in $\Tcurv = \left]-\infty,M\right[$\linebreak (or $\Tcurvb = \left[-\infty,M\right]$).

\subsection{Map of Asplund distances}
\label{sec:AsLIPadd:Map}

\begin{definition}[Map of LIP-additive Asplund distances]
Let $f \in \Fcurv_M$ be a function and $b \in \Tcurv^{D_{b}}$ a probe. The map of Asplund distances is the mapping $Asp_{b}^{\LP}: \Fcurv_M \rightarrow \I$ defined by:
\begin{align}
	Asp_{b}^{\LP} f(x) &= d^{\LP}_{asp} (f_{\left|D_b(x)\right.},b). \label{eq:map_As_add}%
\end{align}
\label{def:map_As_add}
\end{definition}
The LIP addition $\LP$ makes the map of distances robust to contrast variations due to exposure-time changes. 

\subsection{A robust to noise version}
\label{ssec:AsLIPadd:rob_noise}

In order to overcome the noise sensitivity of Asplund metric, a neighbourhood $N_{P,d^{\LP}_{asp},\epsilon,p}(f)$ of the function $f$ is defined by replacing in Eq.~\ref{eq:neighbourhood_tolerance_As_mult} the LIP-multiplicat\-ive Asplund metric $d^{\LT}_{asp}$ by the LIP-additive one $d^{\LP}_{asp}$.
The noise pixels which are the closest to the probe are selected by using a function\linebreak $\gamma^{\LP}_{(f,g)}: D \rightarrow \left]-\infty,M\right[$ which characterises the (LIP-additive) contrast between the functions $f$ and $g$:
\begin{align}
\!\!\!\!\!\!\forall x \in D,\> \gamma^{\LP}_{(f,g)}(x) \LP g(x) = f(x) &\Leftrightarrow \gamma^{\LP}_{(f,g)} = f \LM g.\label{eq:LA_contrast_fct}%
\end{align}
$\gamma^{\LP}_{(f,g)}(x) \in \left]-\infty,M\right[$ is the real value to be LIP-added to each function value $f(x)$ so that it becomes equal to the probe value $g(x)$. 
The expressions of the probes $c_1 \LP g$ and $c_2 \LP g$ (Def. \ref{def:dasAdd}) are equal to: $c_1 = \inf{\{c, \forall x, \gamma^{\LP}_{(f,g)}(x) \leq c \}}$ and $c_2 = \sup{\{c, \forall x, c \leq \gamma^{\LP}_{(f,g)}(x)\}}$. These probes are used to introduce the following definition.

\begin{definition}[LIP-additive Asplund metric with tolerance]
Let $(1-p)$ be a percentage of points of $D$ to be discarded and $D'$ the set of these discarded points. The LIP-additive Asplund metric with tolerance between two functions $f$ and $g \in \Fcurv_M$ is defined by:
\begin{align}
d^{\LP}_{asp,p} (f,g) &= c_{1}' \LM c_{2}'. \label{eq:dAsAdd_tol}%
\end{align}
The constants $c_{1}'$ and $c_{2}'$ are equal to:\linebreak $c_{1}' = \inf \{c, \forall x \in D, \gamma^{\LP}_{(f_{|D \setminus D'},g_{|D \setminus D'})}(x) \leq c \}$ and\linebreak $c_{2}'  = \sup \{ c , \forall x \in D, c \leq \gamma^{\LP}_{(f_{|D \setminus D'},g_{|D \setminus D'})}(x) \}$. A percentage $(1-p)/2$ of the points $x\in D$ with the greatest, respectively lowest contrast values\linebreak $\gamma^{\LP}_{(f,g)}(x) = f(x) \LM g(x)$ are discarded.
\label{def:dAsAdd_tol}
\end{definition}

From Eq.~\ref{eq:LA_contrast_fct} and \ref{eq:LIP:minus}, the contrast function $\gamma^{\LP}_{(f,g)}$ is related to the contrast function $\gamma^{\LP}_{(f,c \LP g)}$, between $f$ and the probe $c \LP g$, where $\forall x \in D, c(x)=c, c \in \left]-\infty,M\right[$, by the following equation: $\gamma^{\LP}_{(f,c \LP g)}(x) = \gamma^{\LP}_{(f,g)}(x) \LM c$. This leads to the following property demonstrated in the supplementary materials.

\begin{property}
\label{prop:inv_LIPAdd_dAsAdd_tol}
The metric $d^{\LP}_{asp,p}$ is invariant under the LIP-addition of a constant.
\end{property}

\begin{remark}
A map of Asplund distances with tolerance can be introduced as in definition \ref{def:map_As_add}:\linebreak $Asp^{\LP}_{b,p}f(x) = d^{\LP}_{asp,p} (f_{|D_b(x)},b)$.
\end{remark}

\begin{figure}[!t]
\centering
\includegraphics[width=0.49\columnwidth]{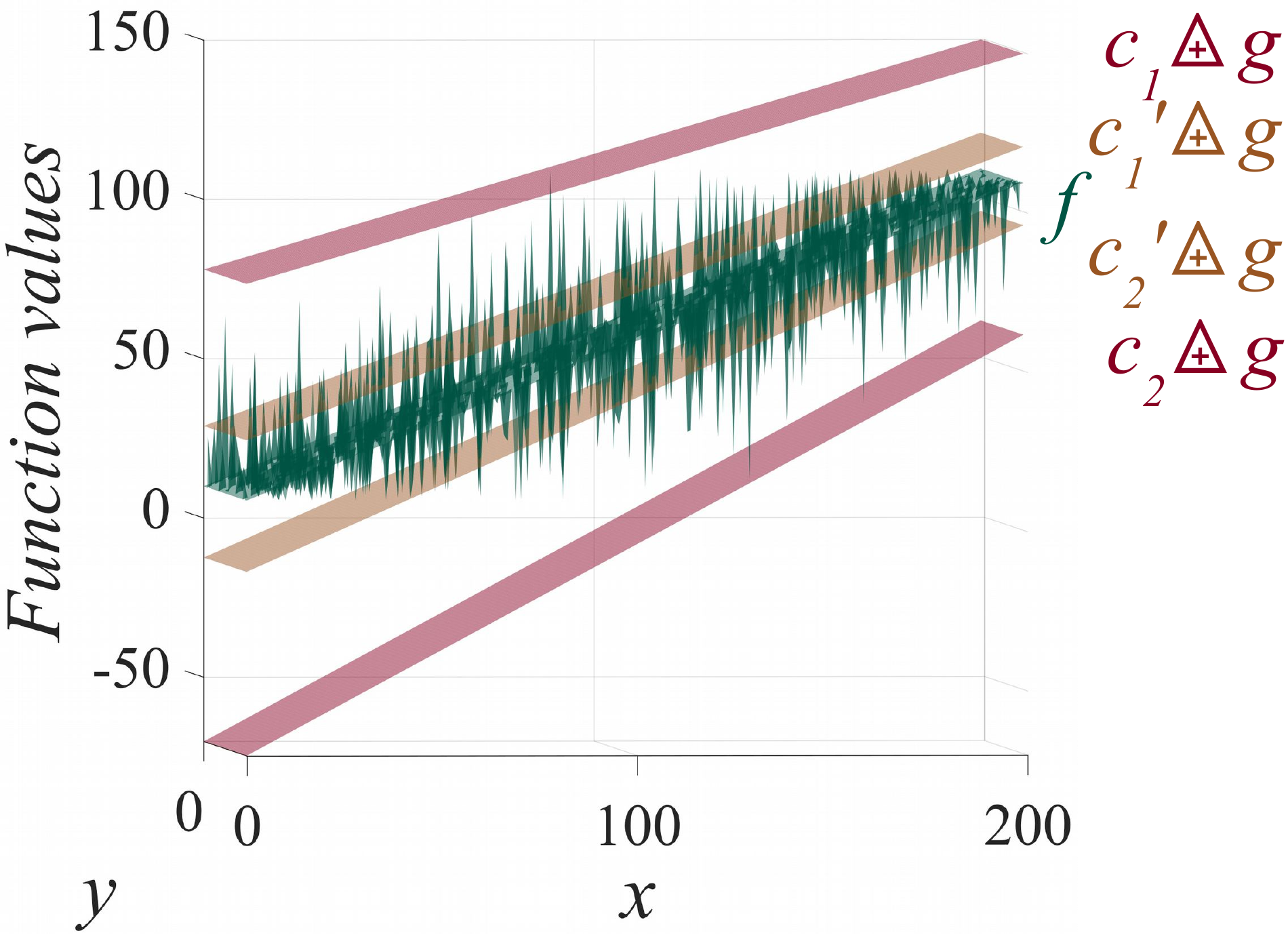}%
\caption{Illustration of the LIP-additive Asplund metric for functions where $g$ is used to probe $f$. $c_1 {\protect \LP} g$ (respectively $c_1' {\protect \LP} g$) is the upper probe and $c_2 {\protect \LP} g$ (resp.  $c_2' {\protect \LP} g$) is the lower probe of the Asplund metric $d^{\protect \LP}_{asp}(f,g)$ (resp. of the Asplund metric with a tolerance $p$, $d^{\protect \LP}_{asp,p}(f,g)$). The function $f$ is obtained by adding a Gaussian white noise to the planar probe $g$ (mean 0, variance 5, spatial density $\rho = 8\%$).} 
\label{fig_AsplundAddTol}
\end{figure}

Fig. \ref{fig_AsplundAddTol} illustrates the LIP-additive Asplund metric robust to noise $d^{\LP}_{asp,p}(f,g)$. The probe $g$ is chosen as a plane and the function $f$ is obtained by adding a noise to the probe $g$ with a spatial density of $\rho = 8\%$. We notice that the upper probe $c_1 \LP g$ and the lower probe $c_2 \LP g$ may take negative values. After the LIP-addition, the planar surface of the probe $g$ is still a plane, with a different orientation. The Asplund distance which is initially equal to\linebreak $d^{\LP}_{asp}(f,g) = 116$ decreases to $d^{\LP}_{asp,p}(f,g) = 40$, with a tolerance $p=97 \%$. Experiments have shown that with $p=90\%$ ($\leq (1-\rho)$), the Asplund distance is equal to zero.

\subsection{Link with Mathematical Morpho-logy}
\label{sec:AsLIPadd:MM}


\subsubsection{General expression for the map of Asplund distances with operations on neighbourhoods}

In order to establish the link with MM, we will express the map of Asplund distances with neighbourhood operations.
From Eq. \ref{eq:dasAdd}, for each $x\in D$, the map expression becomes $d^{\LP}_{asp} (f_{\left|D_b(x)\right.},b) = c_{1_b}f(x) \LIPminus c_{2_b}f(x)$, where $c_{1_b}f(x) =$ $\inf \left\{c, f_{\left|D_b(x)\right.} \leq c \LP b \right\}$ and $c_{2_b} f(x) = \sup \left\{c, c \LP b \leq f_{\left|D_b(x)\right.}\right\}$. This leads to the following definition.

\begin{definition}[LIP-additive maps of the least upper and of the greatest lower bounds]
Let $f \in \Fcurvb_M$ be a function and $b \in \Tcurv^{D_{b}}$ a probe. Their map of the least upper bounds (mlub) $c_{1_b}: \Fcurvb_M  \rightarrow \Fcurvb_M$ and their map of the greatest lower bounds (mglb) $c_{2_{b}}: \Fcurvb_M \rightarrow \Fcurvb_M$ are defined by:
\begin{align}
	c_{1_{b}} f(x) &=  \inf_{h \in D_b}{ \{c, f(x+h) \leq c \LP b(h) \} },  \label{eq:upper_map_add}\\
	c_{2_{b}} f(x) &=  \sup_{h \in D_b}{ \{c, c \LP b(h)	\leq f(x+h) \} }.\label{eq:lower_map_add}%
\end{align}
\end{definition}



The following propositions, \ref{prop:gen_expr_lower_upper_maps_add} to \ref{prop:isomorph_LIPminus}, are demonstrated in the Appendix. The general expressions of the mlub, mglb and map of distances will be given hereinafter.


\begin{proposition}
	The mlub $c_{1_{b}}$ and the mglb $c_{2_{b}}$ are equal to
	\begin{align}
	c_{1_{b}} f(x) &= \bigvee \left\{ f(x+h) \LIPminus b(h) , h \in D_b \right\},	\label{eq:upper_map_add_2}\\
	c_{2_{b}} f(x) &= \bigwedge \left\{ f(x+h) \LIPminus b(h) , h \in D_b \right\}.\label{eq:lower_map_add_2}%
	\end{align}
	\label{prop:gen_expr_lower_upper_maps_add}
\end{proposition}

\begin{corollary}
	The map of Asplund distances between the function $f$ and the probe $b$ is equal to
	\begin{align}
	Asp_{b}^{\LP}f &= c_{1_{b}} f \LIPminus c_{2_{b}} f. \label{eq:map_As_c1_c2_add}%
	\end{align}
	\label{prop:gen_expr_map_As_add}
\end{corollary}


\subsubsection{Particular case of a flat structuring element}


\begin{proposition}
	Let $b=b_0 \in \Tcurv^{D_b}$ be a flat structuring element ($\forall x \in D_b$, $b(x) = b_0$). The mlub $c_{1_{b_0}}$, mglb $c_{2_{b_0}}$ and map of Asplund distances $Asp^{\LP}_{b_0}$ are equal to:
	\begin{align}
		c_{1_{b_0}} f &= \delta_{\overline{D}_b} f \LIPminus b_0 , 												\label{eq:upper_map_add_flat_se}\\
		c_{2_{b_0}} f &= \varepsilon_{D_b} f \LIPminus b_0 ,														\label{eq:lower_map_add_flat_se}\\
		Asp^{\LP}_{b_0} f &= \delta_{\overline{D}_b} f \LIPminus \varepsilon_{D_b} f.	\label{eq:map_As_c1_c2_add_flat_se}%
	\end{align}
	\label{prop:expr_maps_add_flat_se}
\end{proposition}

With a flat probe $b$, the map of Asplund distances is a LIP difference between a dilation $\delta_{\overline{D}_b}$ and an erosion $\varepsilon_{D_b}$ of the function $f$ by the domain $D_b$ of the structuring element $b$. It is similar to a morphological gradient with a LIP difference.

\subsubsection{General case of a structuring function}


\begin{proposition}
	The mlub $c_{1_b}$ (resp. mglb $c_{2_b}$) is a dilation (resp. an erosion) in the same lattice\linebreak $\Lscr_1 = \Lscr_2 = (\Fcurvb_M , \leq)$.
	\label{prop:property_lower_upper_maps_add}
\end{proposition}

\begin{remark}
In Eq. \ref{eq:map_As_c1_c2_add}, one notices that the map of LIP-additive Asplund distances is a LIP-difference between a dilation and an erosion, which corresponds to the LIP version of the morphological gradient. This similarity shows that the map of Asplund distances acts as an operator of derivation.
\end{remark}

Let us establish the relation of the dilation $c_{1_b}$ or the erosion $c_{2_b}$ with the dilation $\delta_b$ (Eq. \ref{eq:dilate_funct}) or the erosion $\varepsilon_b$ (Eq. \ref{eq:erode_funct}), respectively. 
For this purpose, a bijective mapping (i.e. an isomorphism) is needed between the lattice $\Realb^D$ of $\delta_b f$, or $\varepsilon_b f$, and the lattice $\Fcurvb_M$ of $c_{1_b} f$, or $c_{2_b} f$. \citet{Jourlin1995,Navarro2013} defined this isomorphism $\xi: \Fcurvb_M \rightarrow \Realb^D$ and its inverse $\xi^{-1}$ by:
\begin{align}
	\xi(f) 			&= 	-M \ln{(1-f/M)}, 		\label{eq:isomorph_inv_LIP_add}\\
	\xi^{-1}(f) &= 	M(1-\exp{(-f/M)}). \label{eq:isomorph_LIP_add}%
\end{align}

 
\begin{remark} 
As $\xi^{-1}$ and $\xi$ are increasing bijections, they distribute over infima, as well as over suprema. 
\end{remark}


\begin{proposition} Given two functions $f,g \in \Fcurvb_M$, $\xi$ has the property to transform the LIP-difference, $\LM$, into the usual difference, $-$, $\xi(f \LM g) = \xi(f) - \xi(g)$.
	\label{prop:isomorph_LIPminus}
\end{proposition}

The dilation $c_{1_b}$ (Eq.~\ref{eq:upper_map_add_2}) and the erosion $c_{2_b}$ (Eq.~\ref{eq:lower_map_add_2}) can therefore be expressed as:
\begin{align}
	c_{1_b}f(x) &= \xi^{-1} \circ \xi (\bigvee_{h \in \overline{D}_b} \left\{ f(x-h) \LIPminus \overline{b}(h) \right\} \nonumber\\
	&= \xi^{-1} [ \bigvee_{h \in \overline{D}_b} \{ \xi(f)(x-h) - \xi(\overline{b})(h) \} ] \nonumber\\ 
	&= M(1 - e^{-\bigvee_{h \in \overline{D}_b} \{ -\tilde{f}(x-h) + \tilde{\overline{b}}(h) \}} ), \label{eq:upper_map_LIPadd_morpho_ini2}\\
	c_{2_b}f(x) &= \xi^{-1} [ \bigwedge_{h \in D_b} \{ \xi(f)(x+h) - \xi(b)(h) \} ] \nonumber\\ 
	&= M(1 - e^{-\bigwedge_{h \in D_b} \{ -\tilde{f}(x+h)) - [-\tilde{b}(h)] \} }), \label{eq:lower_map_LIPadd_morpho_ini2}%
\end{align}
where $\xi(f) = -M \tilde{f}$.

\begin{proposition}
	Let $b \in \Tcurv^{D_b}$ be a structuring function and $f \in \Fcurvb_M$ be a function. The expressions of their mlub $c_{1_b}$, which is a dilation, and of their mglb $c_{2_b}$, which is an erosion, are equal to:
	\begin{align}
		c_{1_b} f &= \xi^{-1} [ \xi(f) \oplus (-\xi(\overline{b})) ]\nonumber\\
							&= M (1 - e^{-[ \acute{f} \oplus (-\acute{\overline{b}})] }), \label{eq:upper_map_LIPadd_morpho}\\
		c_{2_b} f &= \xi^{-1} [ \xi(f) \ominus \xi(b)  ]\nonumber\\
							&= M (1 - e^{-[ \acute{f} \ominus \acute{b} ] }). 					 \label{eq:lower_map_LIPadd_morpho}%
	\end{align}
	$\acute{f} \in \Realb^D$ is a function defined by $\acute{f} = - \tilde{f} = -\ln{\left( 1 - f/M \right)}$ $= [\xi(f)]/M$.
	
	Their map of Asplund distances is related to the difference between a dilation and an erosion (with an additive structuring function) by:
	\begin{align}
	\!\!\!\!\!\!\!Asp_{b}^{\LP}f &= \xi^{-1} [ [\xi(f) \oplus (-\xi(\overline{b}))] - [\xi(f) \ominus \xi(b)] ] 			\label{eq:AsAdd_map_morpho2}\\
		&= M (1 - \exp{(-[ (\acute{f} \oplus (-\acute{\overline{b}})) - (\acute{f} \ominus \acute{b}) ]) } )		\nonumber\\
		&= M (1 - \exp{(-[ \delta_{-\acute{\overline{b}}} \acute{f} - \varepsilon_{\acute{b}} \acute{f} ] )} ).	\label{eq:AsAdd_map_morpho}
	\end{align}
	\label{prop:expr_mapsAdd_morpho}
\end{proposition}

\begin{remark} By replacing the dilation and erosion by rank-filters \citep{Serra1988} one can compute the map of Asplund distances with a tolerance $Asp_{b,p}^{\LP}$.
\end{remark}

Fig. \ref{fig_MapAsplundAdd_signal:1} illustrates the double-sided probing of an image $f$ by a probe (or structuring function) $b$.
The amplitudes of both peaks are related by a LIP-addition of a constant. In Fig. \ref{fig_MapAsplundAdd_signal:2}, the two deepest minima of the map of distances of $f$, $As_b^{\LP} f$ correspond to the location of both peaks similar to the probe $b$. This illustrates the insensitivity of the map of LIP-additive Asplund distances to a variation of light intensity or exposure-time.\columnbreak

\begin{figure}[!htbp]
\centering
\subfloat[]{\includegraphics[width=0.49\columnwidth]{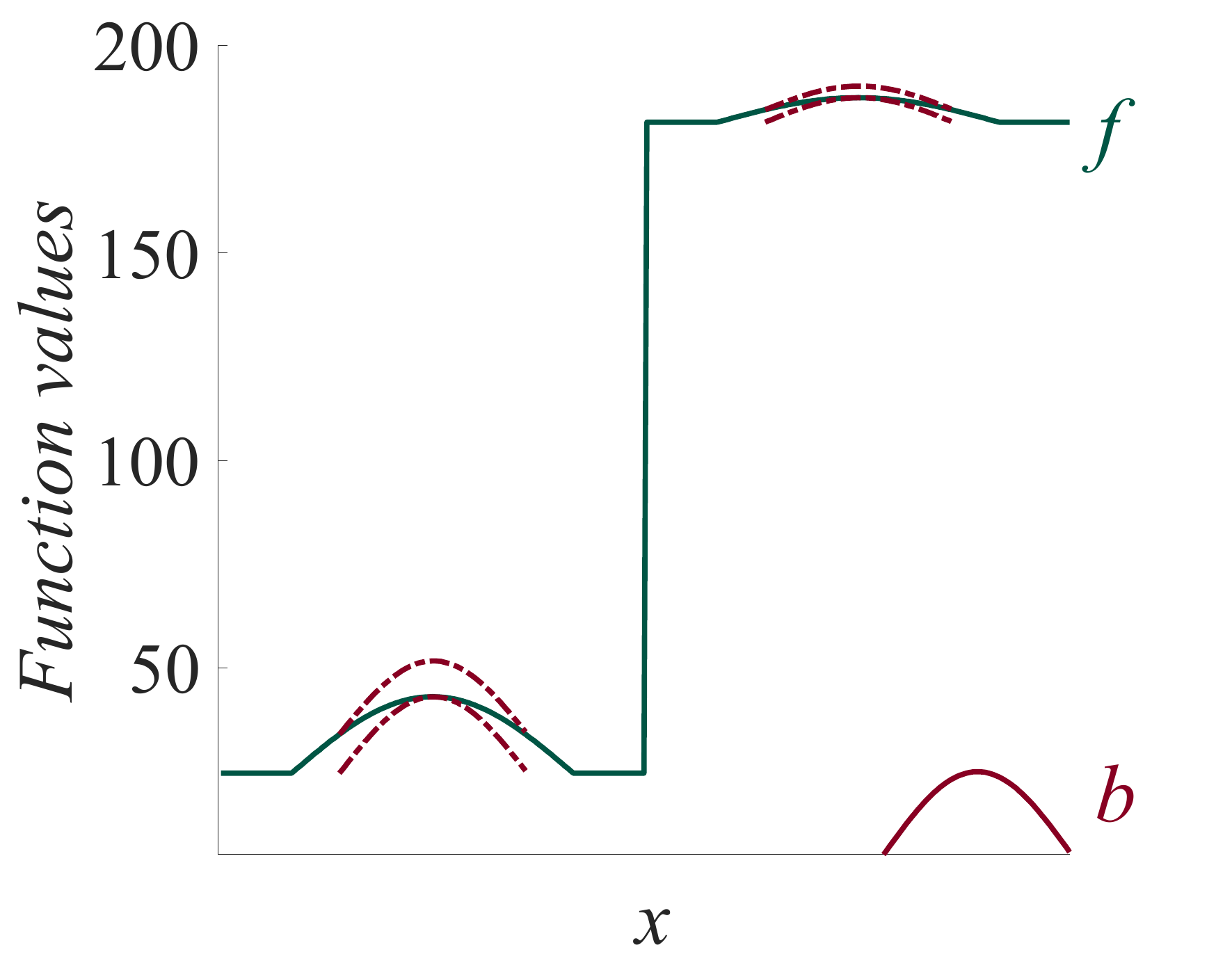}%
\label{fig_MapAsplundAdd_signal:1}}
\hfil
\subfloat[]{\includegraphics[width=0.49\columnwidth]{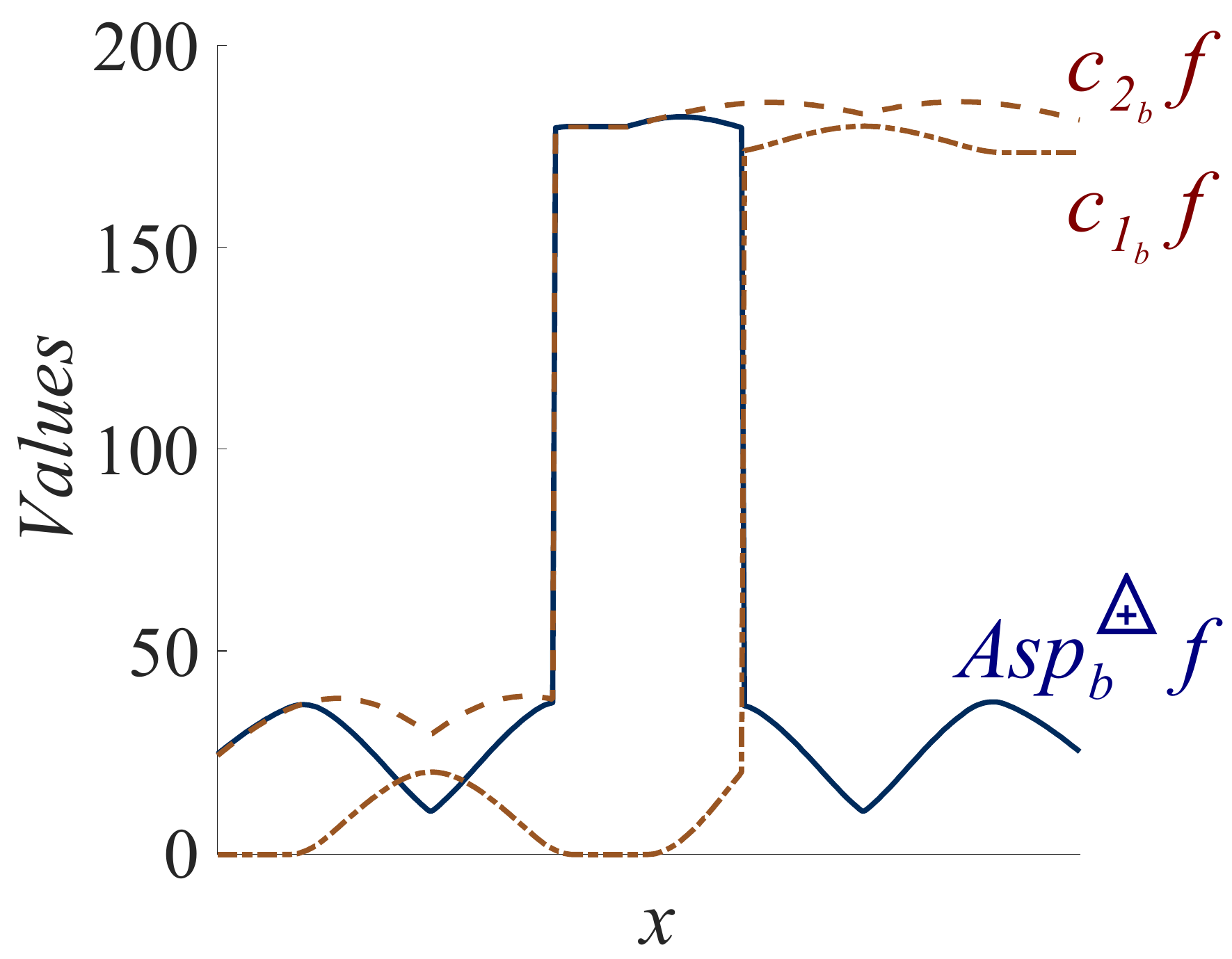}%
\label{fig_MapAsplundAdd_signal:2}}
\caption{(a) The function $f$ is probed on both sides (up and down) by the probe $b$ (in red) using the LIP-additive law ${\protect \LP}$. The double-probing is depicted at two locations by the red dashed curves. (b) $c_{1_b} f$ and $c_{2_b} f$ are the mlub and the mglb of $f$, respectively. $Asp_{b}^{\protect \LP} f$ is the map of Asplund distances between $f$ and $b$.}
\label{fig_MapAsplundAdd_signal}
\end{figure}


\section{Experiments and results}
\label{sec:illus}

In this section, we will discuss the implementation of the maps of distances and we will illustrate them with simulated and real cases of image acquisition.

\subsection{Implementation}

The maps of LIP-multiplicative and LIP-additive Asplund distances of the image $f$, $Asp^{\LT}_b f$ and $Asp^{\LP}_b f$, respectively, were both programmed in MATLAB\textsuperscript{\tiny\textregistered} using their direct (Eq.~\ref{eq:map_AsMult}, \ref{eq:map_As_add}) and morphological (Eq.~\ref{eq:As_mapMult_morpho}, \ref{eq:AsAdd_map_morpho}) expressions. The same programming was done for the map of Asplund distances with tolerance $Asp^{\LT}_{b,p} f$ and $Asp^{\LP}_{b,p} f$ of $f$. Due to the existence in numerous image analysis software (e.g. MATLAB\textsuperscript{\tiny\textregistered}, Python scikit-images) of fast versions of the morphological operations $\oplus$ and $\ominus$ and of rank filters, the morphological implementation is easier and faster than the direct one. In table \ref{tab:Simulated_Christmas_balls:Duration}, the duration of the morphological and of the direct implementations (with parallelisation) of the maps of Asplund distances without and with tolerance are compared using the example of Fig. \ref{fig:Simulated_Christmas_calendar}. The image size is of 1224~$\times$~918 pixels and the probe contains 285 pixels (Fig.~\ref{fig:Simulated_Christmas_calendar:4}). The morphological implementation is always faster than the direct one, with a gain factor lying between 5.9 and 11 (processor Intel\textsuperscript{\tiny\textregistered} Core\textsuperscript{\tiny{TM}} i7 CPU 4702HQ, 2.20~GHz, 4 cores, 8 threads with 16~Gb RAM).\pagebreak

\begin{table}[!t]
\caption{Comparison between the durations of the morphological and direct implementations for the LIP-multiplicative maps of Asplund distances, $Asp^{\protect \LT}_b$ and $Asp^{\protect \LT}_{b,95\%}$, and the LIP-additive maps, $Asp^{\protect \LP}_{b}$ and $Asp^{\protect \LP}_{b,94\%}$. The example of Fig. \ref{fig:Simulated_Christmas_calendar} is used. Last raw: gain factor between the direct and the morphological implementations.}
\label{tab:Simulated_Christmas_balls:Duration} 
\centering
\renewcommand*{\arraystretch}{1.7}
\begin{tabular}{@{}c@{\hspace{0.5em}}|@{\hspace{0.5em}}c@{\hspace{0.5em}}|@{ }c@{ }|@{\hspace{0.5em}}c@{\hspace{0.5em}}|@{\hspace{0.5em}}c@{}}
\hline
									& $Asp^{\protect \LT}_b f$ 	& $Asp^{\protect \LT}_{b,95\%}f$		& $Asp^{\protect \LP}_{b}f$ 		& $Asp^{\protect \LP}_{b,94\%}f$ 	\\
\hline
Morpho. 		 & \hphantom{0}1.4\rlap{ s}  	& \hphantom{0}4.2\rlap{ s} 					& \hphantom{0}1.4\rlap{ s} 		&  \hphantom{0}4.3\rlap{ s}\\
\hline
Direct         		 & 16.1\rlap{ s} 					 		& 24.6\rlap{ s} 				 	 					& 15.4\rlap{ s} 							&  25.5\rlap{ s} 					 \\
\hline
Gain factor				 & 10.9   					 					& \hphantom{0}5.9   			  				& 11.0  				  						&  \hphantom{0}5.9   			 \\
\hline
\end{tabular}
\end{table}

\subsection{Simulated cases}

We evaluate the LIP-multiplicative and LIP-additive maps to detect balls in a bright image and in two darkened versions obtained by simulation (Fig.~\ref{fig:Simulated_Christmas_calendar}). The bright image is acquired in colour with ``normal'' contrasts -- automatically selected by the camera (Fig.~\ref{fig:Simulated_Christmas_calendar:1}) -- and converted to a luminance  image in grey-level, denoted $f$. A first darkened version $f_{dk,\LT}$ is obtained by the LIP-multiplication of $f$ by a scalar $\la$ (Fig.~\ref{fig:Simulated_Christmas_calendar:2}). A second darkened version $f_{dk,\LP}$ is obtained by the LIP-addition of a constant $k$ (Fig.~\ref{fig:Simulated_Christmas_calendar:5}). 
Due to the light reflection at the surface of the balls and of other confounding objects, detecting the balls is a difficult task in these images. For this purpose, we used a probe $b$ made of a ring surrounding a cylinder (Fig.~\ref{fig:Simulated_Christmas_calendar:4}). The ring has an external radius of 15 pixels, a width of 3 pixels, a grey-level value of 18 and the disk has a radius of 2 pixels and a grey-level value of 190. 

\begin{remark}
As the grey-scale is complemented in the LIP model, the LIP-multiplicative map $As^{\protect \LT}_{b^c,p} f^c$ is computed using the image complement\linebreak $f^c=M-f \in \left]0,M\right[^D$, and the probe complement $b^c$. However, the image $f_{dk,\LP} \in \left]-\infty,M\right[^D$, darkened by the LIP-addition of a constant, presents negative values and its complement $f_{dk,\LP}^c \in \left]0,+\infty \right[^D$ is outside of the dynamic range allowed for the map $\left[-\infty,M\right]^D$. For this reason, the darkened image $f_{dk,\LP}$ is not complemented.
\end{remark}

In the image $f$ (Fig.~\ref{fig:Simulated_Christmas_calendar:1}) and in its darkened version $f_{dk,\LT}$ (Fig.~\ref{fig:Simulated_Christmas_calendar:2}) obtained by LIP-multiplication, all the balls are detected by a thresholding of their map of Asplund distances $Asp^{\LT}_{b^c,p} f^c$ (Fig.~\ref{fig:Simulated_Christmas_calendar:3}). The map is the same for both images $f$ and $f_{dk,\LT}$ because of its invariance under LIP-multiplication by a scalar. Similar results are obtained for the image $f$ (Fig.~\ref{fig:Simulated_Christmas_calendar:1}), its darkened version $f_{dk,\LP}$ by LIP-addition (Fig.~\ref{fig:Simulated_Christmas_calendar:5}) and their map of distances $Asp^{\LP}_{b,p} f$ (Fig.~\ref{fig:Simulated_Christmas_calendar:6}) which is invariant under LIP-addition of a constant. The same parameters of the probe were used for all the images $f$, $f_{dk,\LT}$ and $f_{dk,\LP}$.
These results show that the maps of Asplund distances are able to detect targets under different illumination conditions modelled by the LIP-multiplication $\LT$ or the LIP-addition $\LP$.

\begin{figure}[!htbp]
\centering
\subfloat[]{\includegraphics[width=0.32\columnwidth]{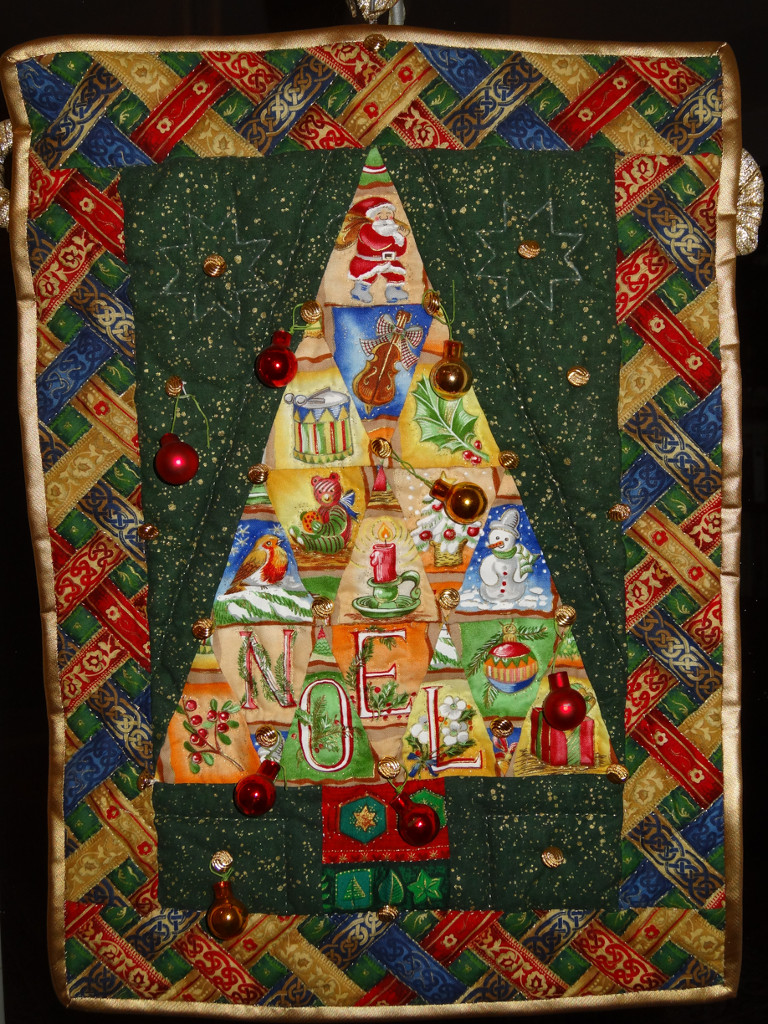}%
\label{fig:Simulated_Christmas_calendar:1}}
\hfil
\subfloat[]{\includegraphics[width=0.32\columnwidth]{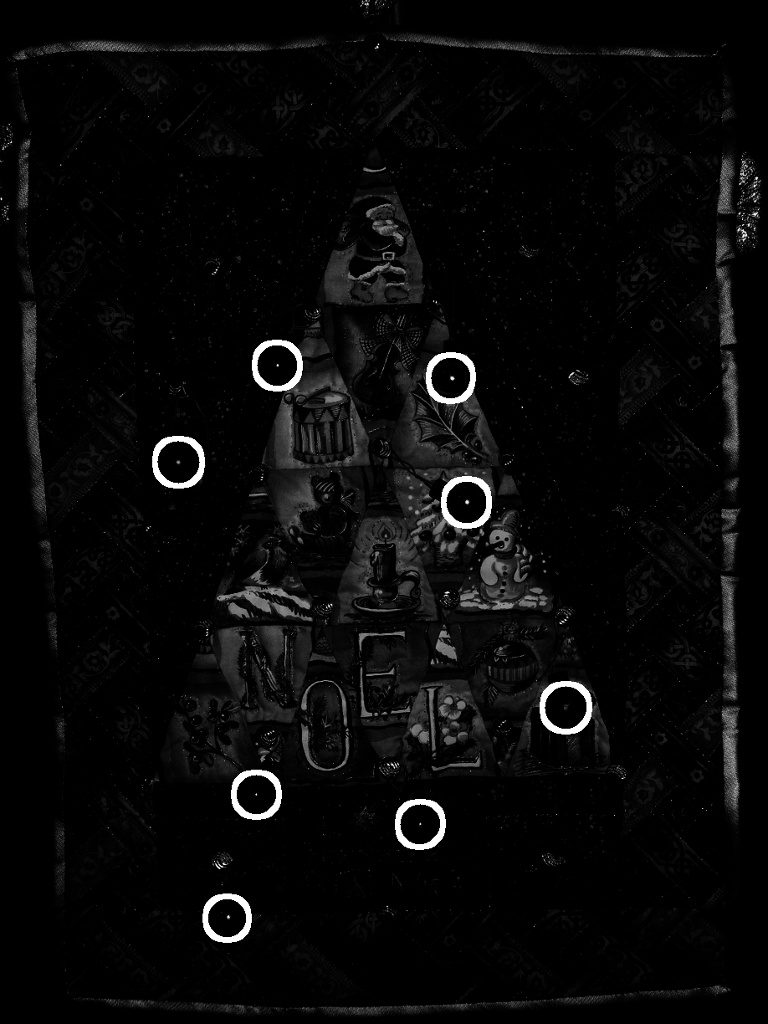}%
\label{fig:Simulated_Christmas_calendar:2}}
\hfil
\subfloat[]{\includegraphics[width=0.32\columnwidth]{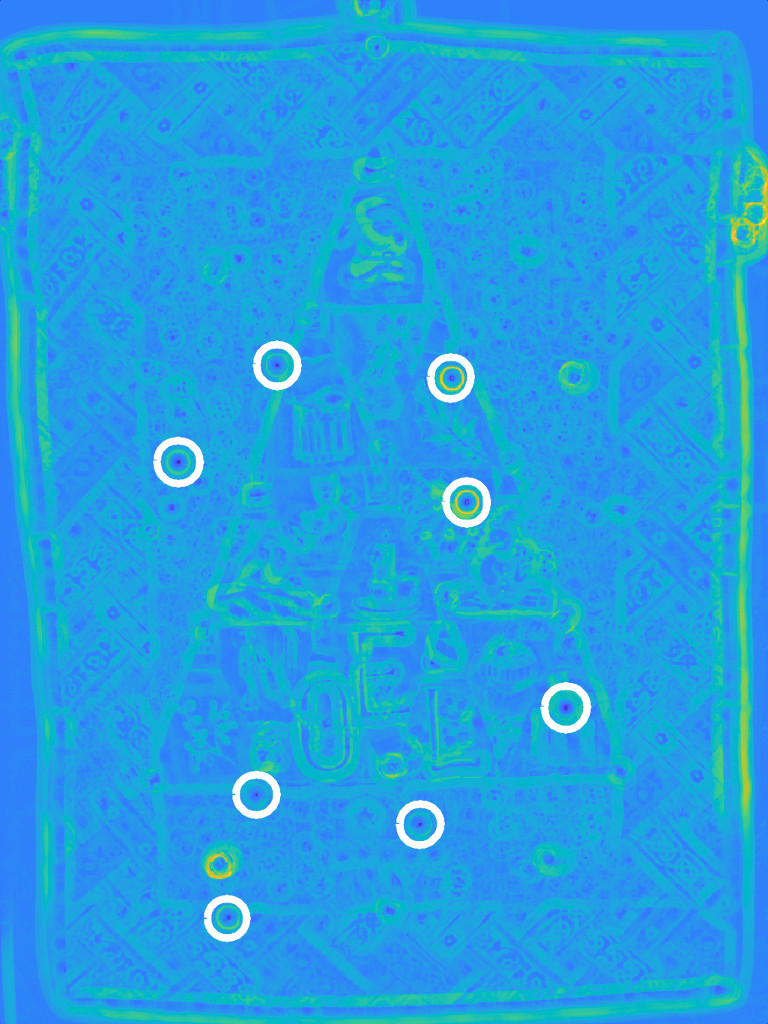}%
\label{fig:Simulated_Christmas_calendar:3}}
\hfil
\subfloat[]{\includegraphics[width=0.32\columnwidth]{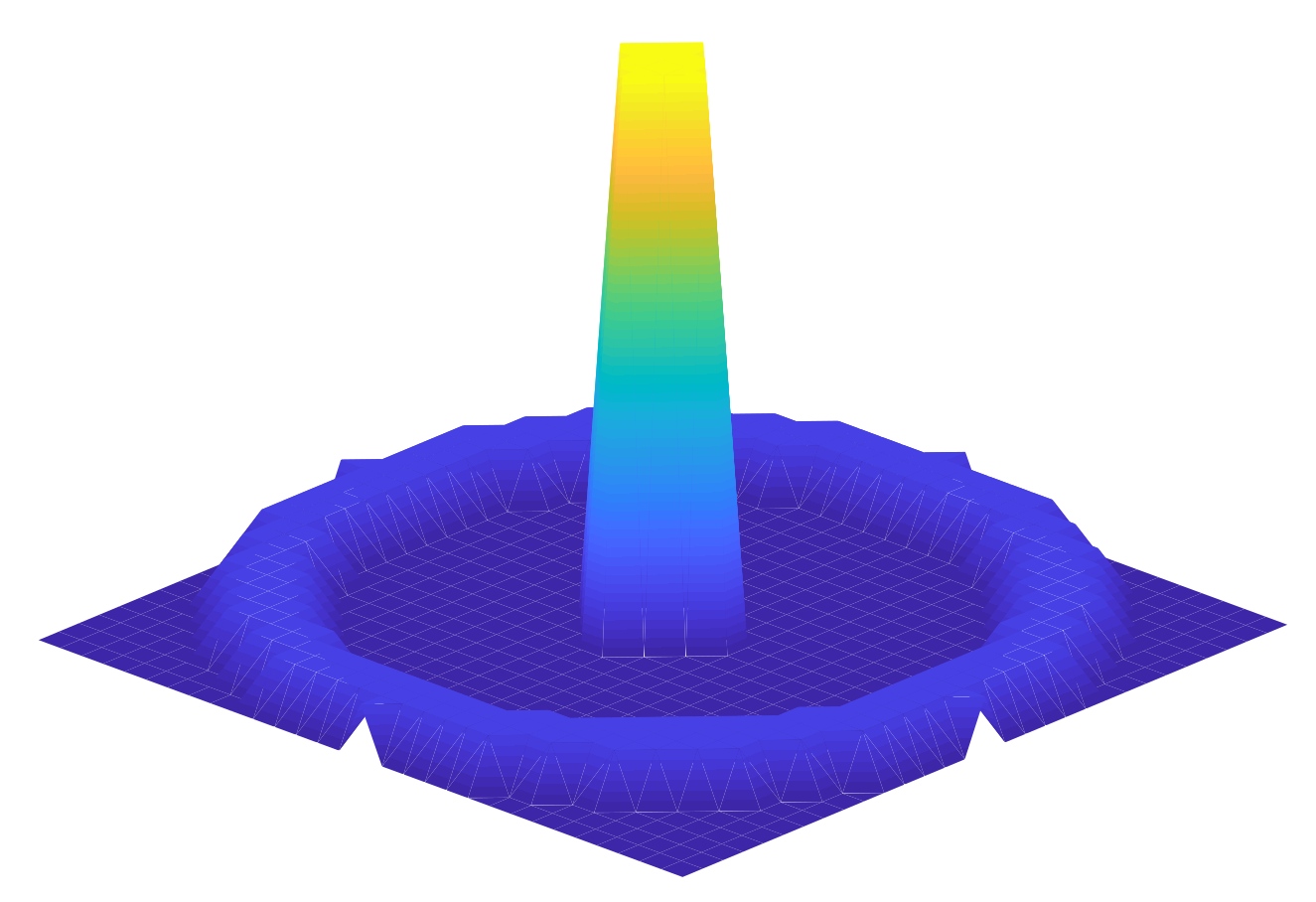}%
\label{fig:Simulated_Christmas_calendar:4}}
\hfil
\subfloat[]{\includegraphics[width=0.32\columnwidth]{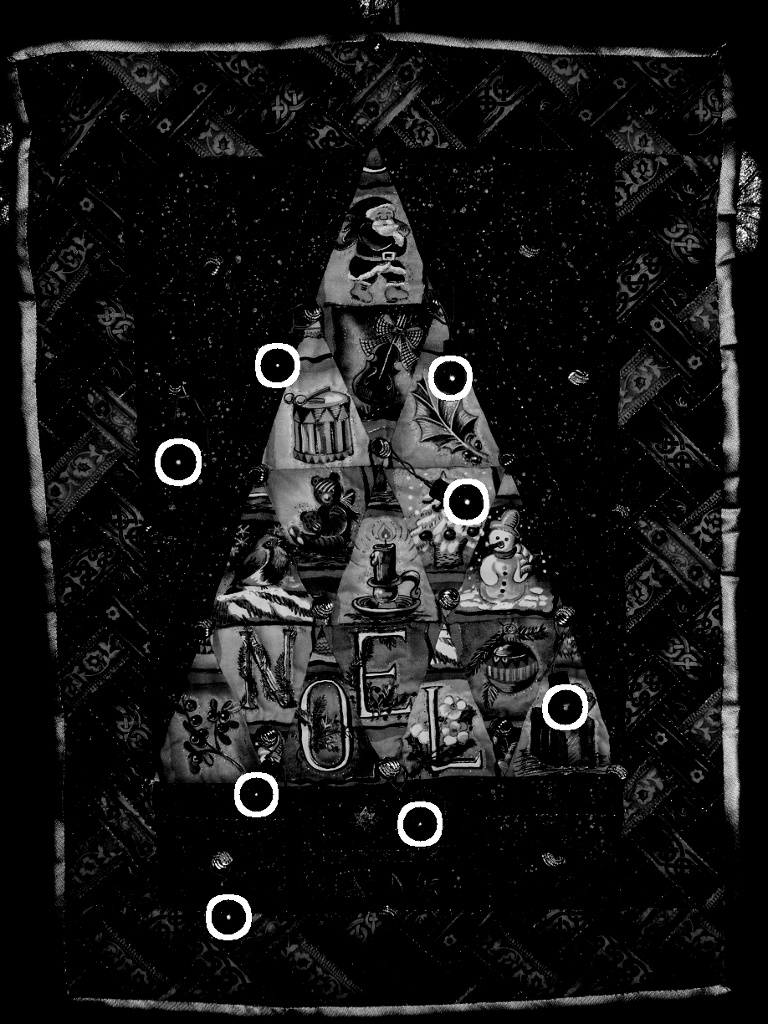}%
\label{fig:Simulated_Christmas_calendar:5}}
\hfil
\subfloat[]{\includegraphics[width=0.32\columnwidth]{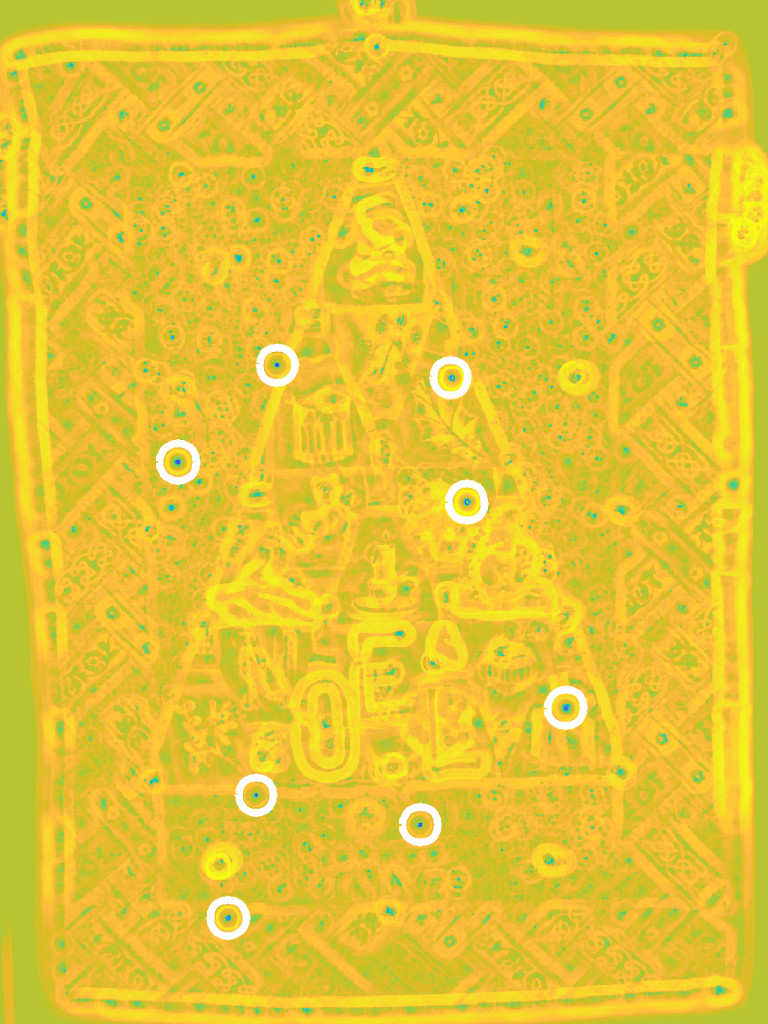}%
\label{fig:Simulated_Christmas_calendar:6}}
\caption{(a) Colour version of the luminance image $f$ and (b) its darkened version $f_{dk,{\protect \LT}} = (\la {\protect \LT} f^c)^c$ obtained by the LIP-multiplication ($\la = 5$) of the complemented image luminance $f^c$. The balls are detected by a thresholding of (c) the map of LIP-multiplicative Asplund distances of $f^c$ with a tolerance $p = 95\%$, $Asp^{\protect \LT}_{b^c,p} f^c$. (d) Probe $b$. (e) Darkened version $f_{dk,{\protect \LP}} = k {\protect \LP} f$ of the luminance image $f$ obtained by the LIP-addition of $k = {\protect \LM} 100 = -164.10$. The balls are segmented by a thresholding of (f) the map of LIP-additive Asplund distances of $f$ with a tolerance $p = 94\%$, $Asp^{\protect \LP}_{b,p} f$.}
\label{fig:Simulated_Christmas_calendar}
\end{figure}

\subsection{Real cases}

The maps of LIP-additive and LIP-multiplicative Asplund distances are then illustrated on real images. 
For a better visual interpretation, the results are presented with colour images even if the processing is made using their luminance.

\subsubsection{LIP-additive metric: images acquired with a variable exposure-time}

In Fig. \ref{fig:Different_exposure_times_Christmas_balls}, the same scene is captured with three different camera exposure-times (or shutter speeds). This gives three images, each with a different brightness: a bright one $f$ (Fig.~\ref{fig:Different_exposure_times_Christmas_balls:1}), a dark intermediate one $f_{dk_1}$ (Fig.~\ref{fig:Different_exposure_times_Christmas_balls:2}) and a dark one $f_{dk_2}$ (Fig.~\ref{fig:Different_exposure_times_Christmas_balls:3}). The scene is composed of bright balls on a multicolour background with other smaller balls acting as confounding objects. In order to make the ball detection more arduous, the camera is not exactly in the same position to capture the images. Moreover, the balls are of different colours, with different backgrounds and present several reflections. A disk of diameter 55 pixels is manually selected inside a ball of the bright image $f$ (Fig.~\ref{fig:Different_exposure_times_Christmas_balls:1}) in order to serve as a probe function $b$. A map of Asplund distances is computed between the complement of each of the three images and the same probe $b$ using the same tolerance parameter $p=70\%$. In the three distance maps of images $Asp^{\protect \LP}_{b^{c},p} f^{c}$ (Fig.~\ref{fig:Different_exposure_times_Christmas_balls:4}), $Asp^{\protect \LP}_{b^{c},p} f_{dk_1}^{c}$ (Fig.~\ref{fig:Different_exposure_times_Christmas_balls:5}) and $Asp^{\protect \LP}_{b^{c},p} f_{dk_2}^{c}$ (Fig.~\ref{fig:Different_exposure_times_Christmas_balls:6}), one can notice that the amplitudes are similar. This is caused by the low sensitivity of the LIP-additive Asplund metric to lighting variations due to different exposure-times. In order to extract the location of the large balls, the maps of the dark intermediate image $Asp^{\protect \LP}_{b^{c},p} f_{dk_1}^{c}$ and of the dark image $Asp^{\protect \LP}_{b^{c},p} f_{dk_2}^{c}$ are segmented using the same technique -- a threshold (at the $37^{th}$ percentile) and a reconstruction of the regional h-minima \citep{Soille2003} -- followed by a morphological post-processing (see remark in supplementary materials). Such a technique allows to detect all the large balls in both dark images $f_{dk_1}$ (Fig.~\ref{fig:Different_exposure_times_Christmas_balls:2}) and $f_{dk_2}$ (Fig.~\ref{fig:Different_exposure_times_Christmas_balls:3}), using the same probe $b$ extracted in the bright image $f$ (Fig.~\ref{fig:Different_exposure_times_Christmas_balls:1}). This illustrates the robustness of the map of LIP-additive Asplund distances to different exposure-times.

\begin{figure}[!htbp]
\centering
\subfloat[]{\includegraphics[width=0.33\columnwidth]{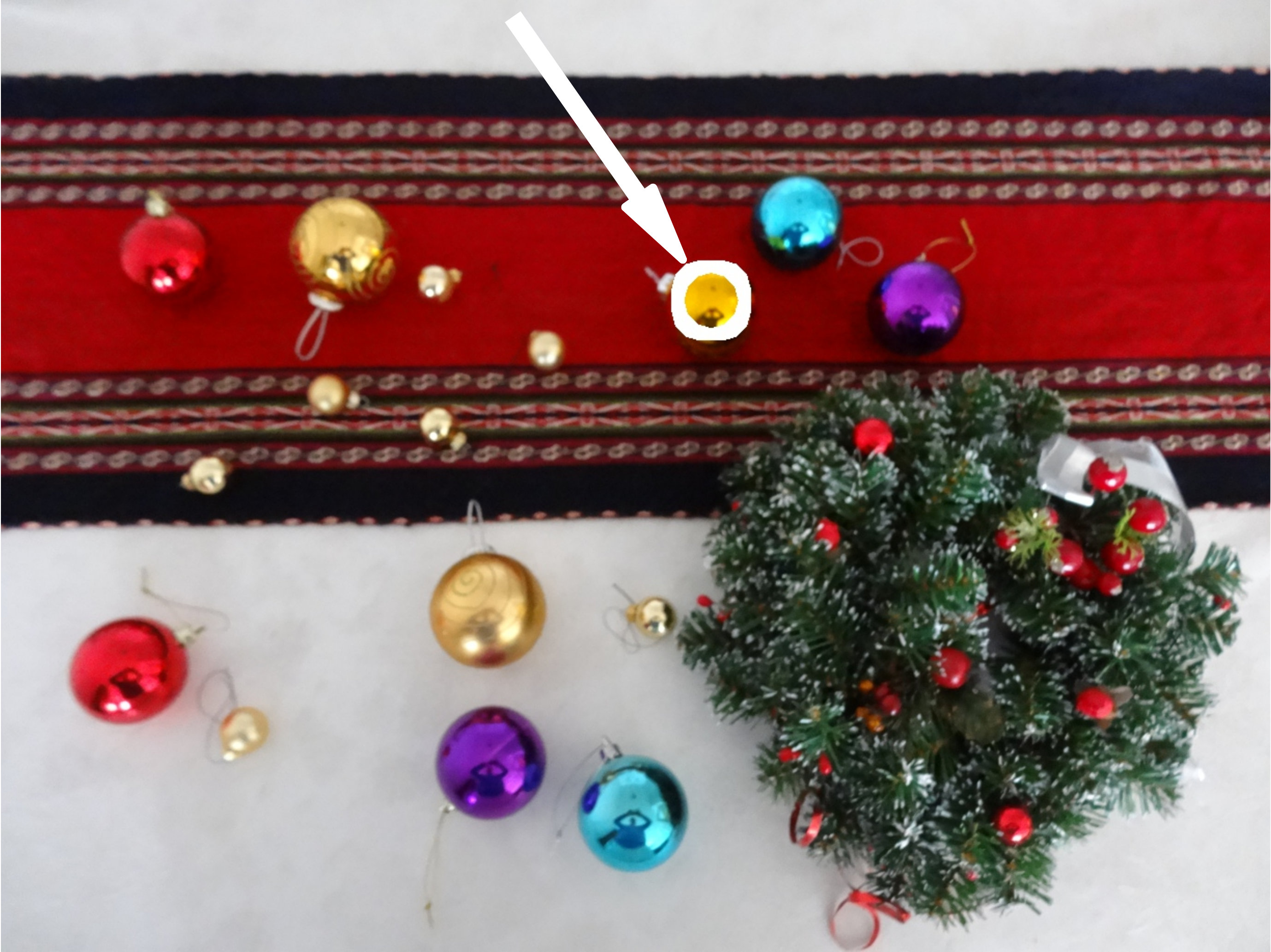}%
\label{fig:Different_exposure_times_Christmas_balls:1}}
\hfil
\subfloat[]{\includegraphics[width=0.33\columnwidth]{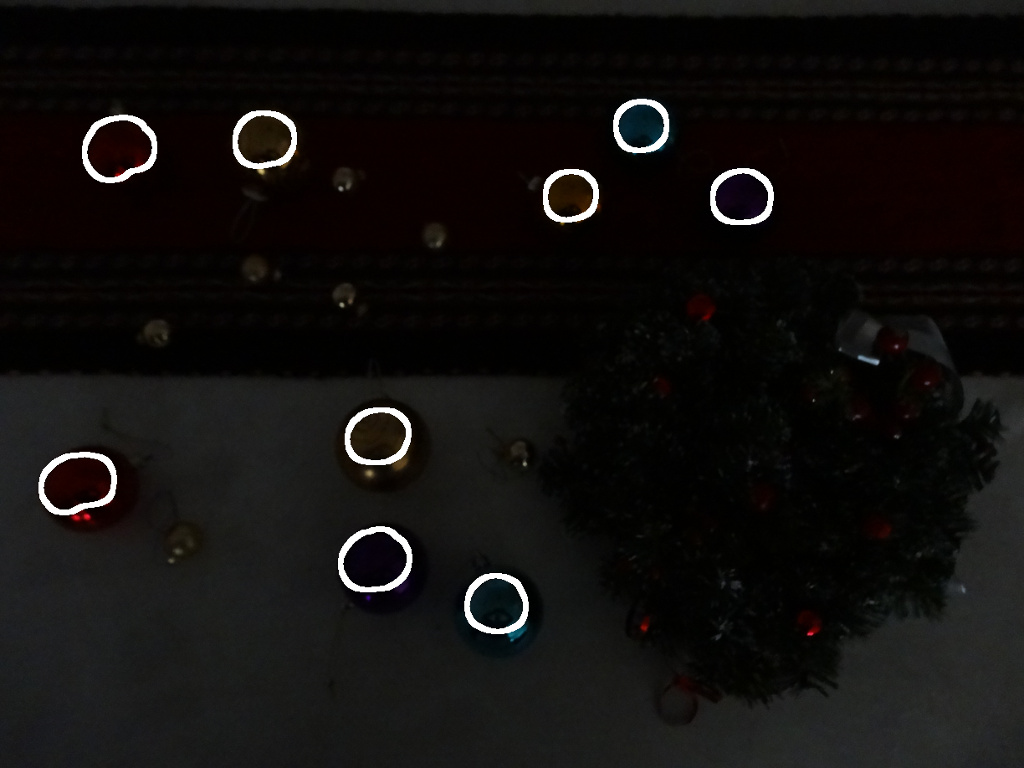}%
\label{fig:Different_exposure_times_Christmas_balls:2}}
\hfil
\subfloat[]{\includegraphics[width=0.33\columnwidth]{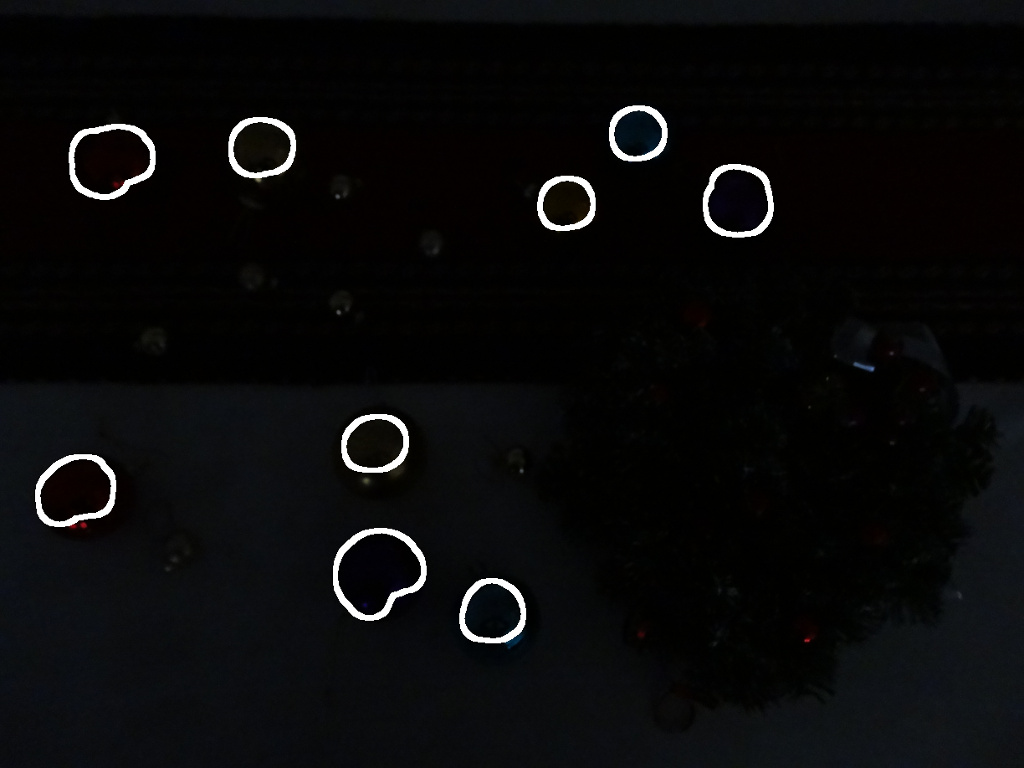}%
\label{fig:Different_exposure_times_Christmas_balls:3}}
\hfil
\subfloat[]{\includegraphics[width=0.33\columnwidth]{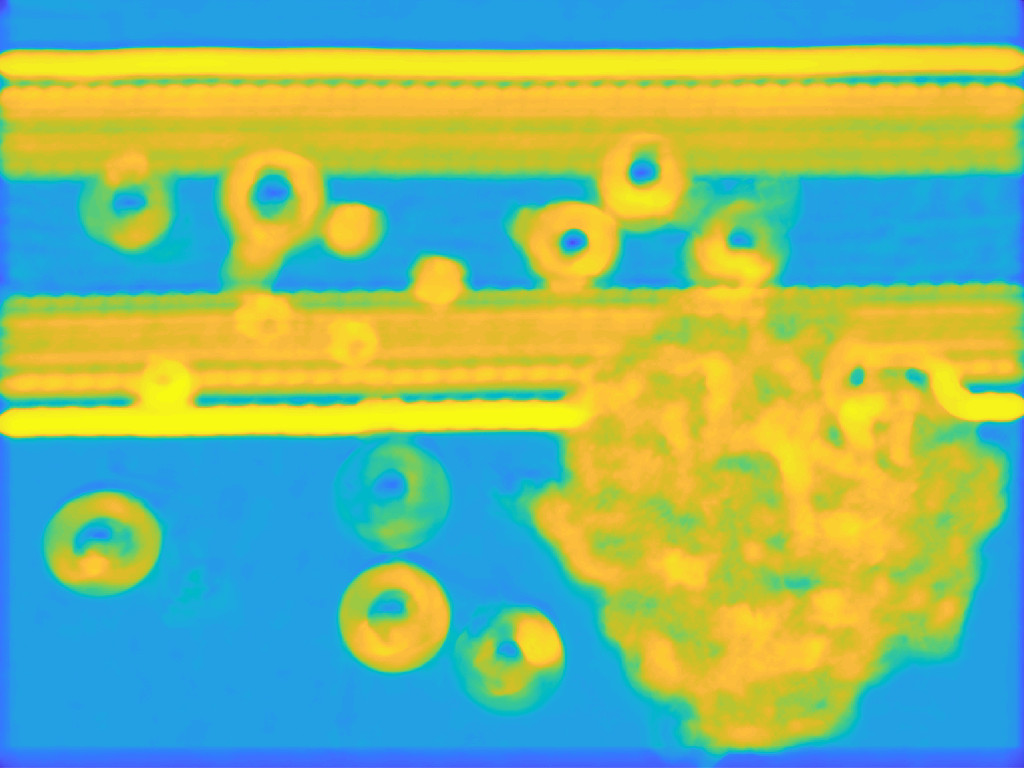}%
\label{fig:Different_exposure_times_Christmas_balls:4}}
\hfil
\subfloat[]{\includegraphics[width=0.33\columnwidth]{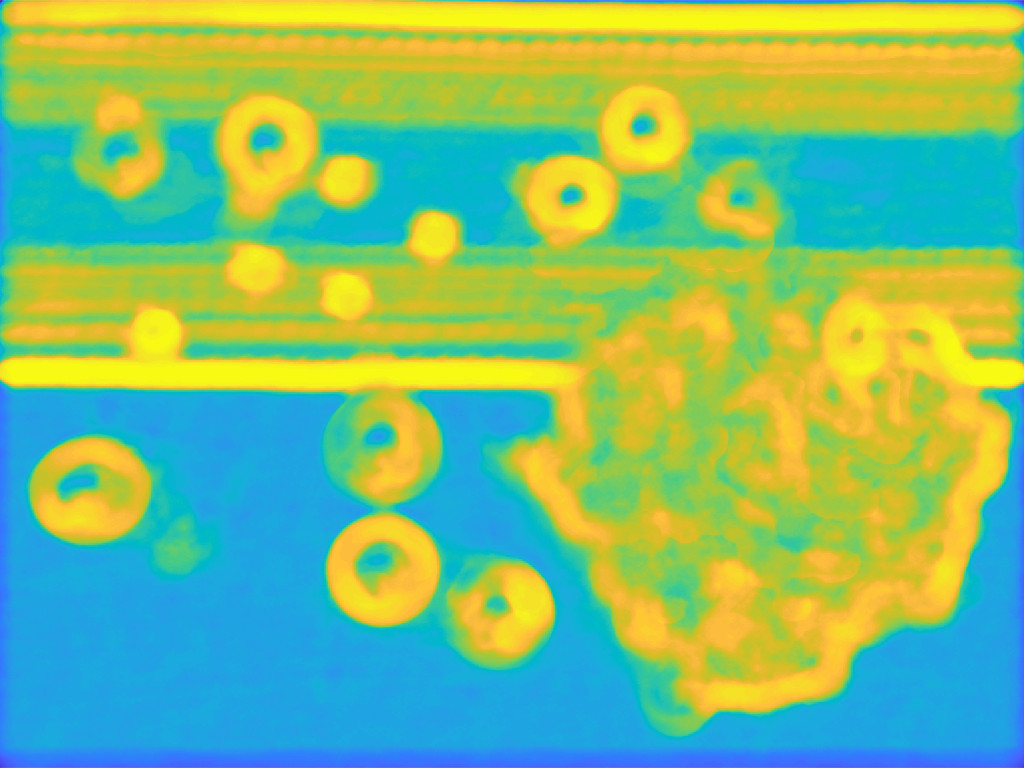}%
\label{fig:Different_exposure_times_Christmas_balls:5}}
\hfil
\subfloat[]{\includegraphics[width=0.33\columnwidth]{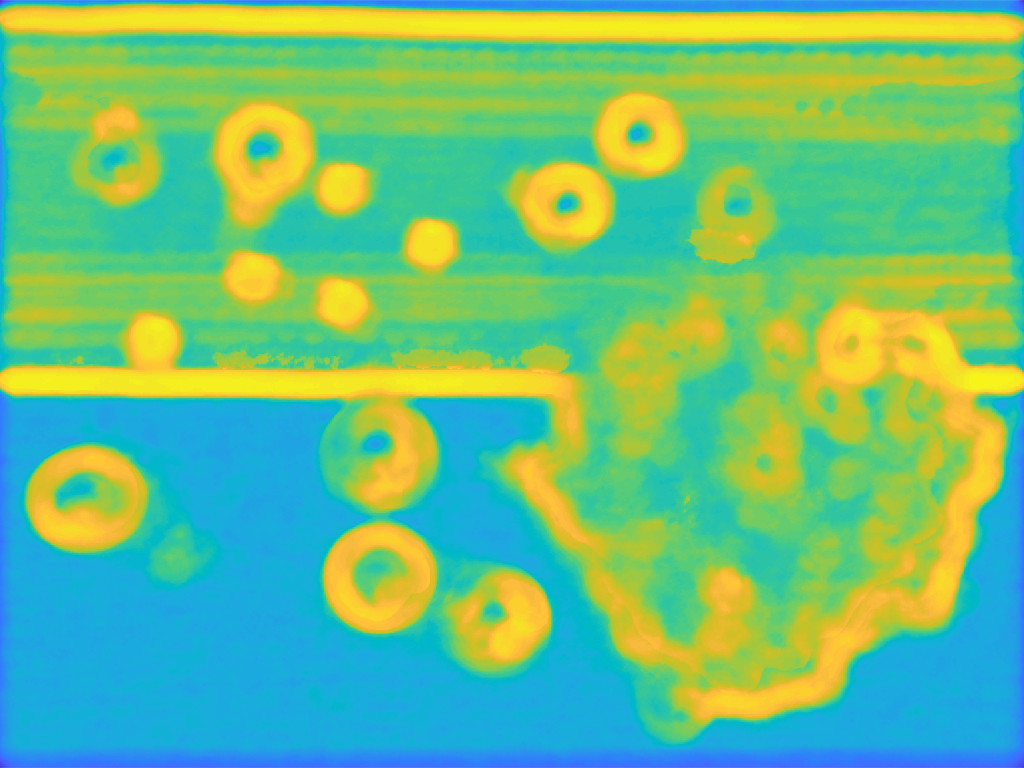}%
\label{fig:Different_exposure_times_Christmas_balls:6}}
\caption{(a) Bright image $f$ acquired with an exposure-time of 1/5~\textnormal{s}. The probe function $b$ is shown by the white arrow. (b) Ball detection in a dark intermediate image $f_{dk_1}$ (exposure-time of 1/80~\textnormal{s}). (c) Ball detection in a dark image $f_{dk_2}$ (exposure-time of 1/160~\textnormal{s}). (d) Map of LIP-additive Asplund distances $Asp^{\protect \LP}_{b^{c},p} f^{c}$ of the image $f$, (e) map $Asp^{\protect \LP}_{b^{c},p} f_{dk_1}^{c}$ of the image $f_{dk_1}$ and (f) map $Asp^{\protect \LP}_{b^{c},p} f_{dk_2}^{c}$ of the image $f_{dk_2}$. The tolerance parameter $p$ is set to 70\%.}
\label{fig:Different_exposure_times_Christmas_balls}
\end{figure}
\columnbreak

\begin{figure}[!t]
\centering
\subfloat[]{\includegraphics[width=0.33\columnwidth]{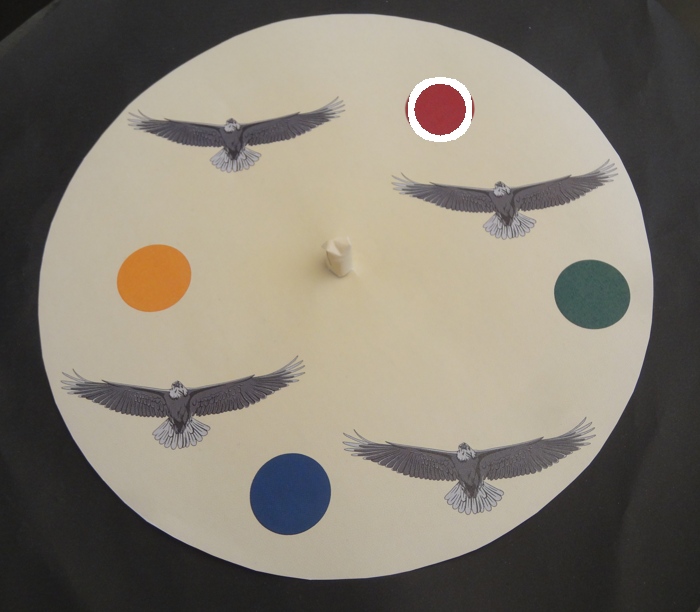}%
\label{fig:Moving_disk:1}}
\hfil
\subfloat[]{\includegraphics[width=0.33\columnwidth]{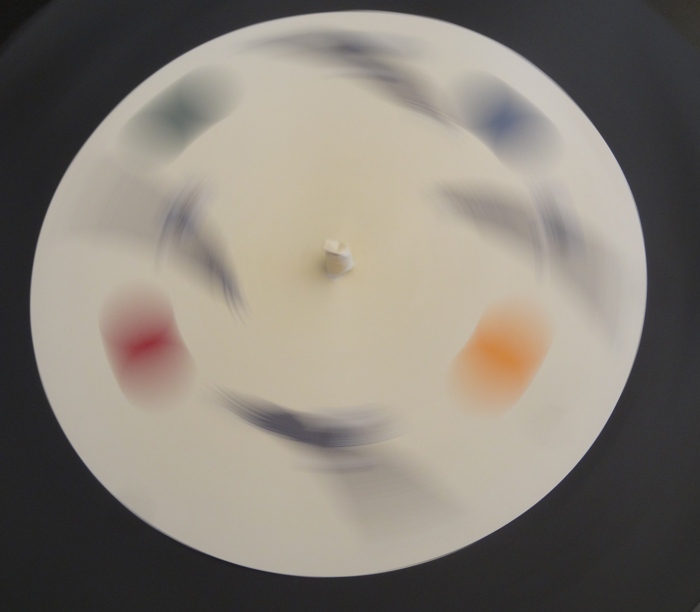}%
\label{fig:Moving_disk:2}}
\hfil
\subfloat[]{\includegraphics[width=0.33\columnwidth]{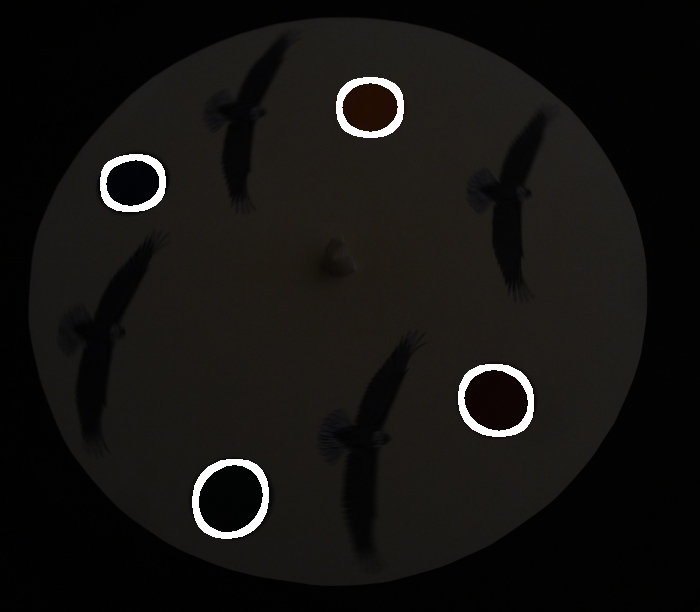}%
\label{fig:Moving_disk:3}}

\subfloat[]{\includegraphics[width=0.33\columnwidth]{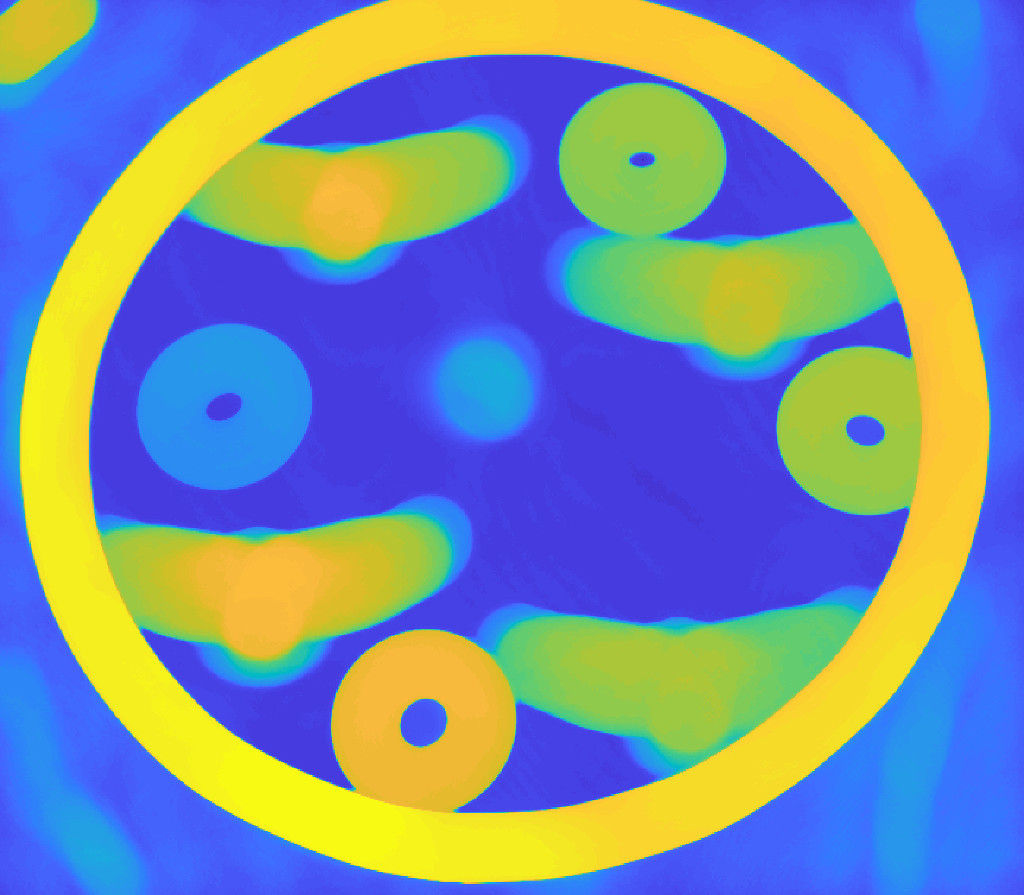}%
\label{fig:Moving_disk:4}}
\hfil
\hspace{0.32\columnwidth}
\hfil
\subfloat[]{\includegraphics[width=0.33\columnwidth]{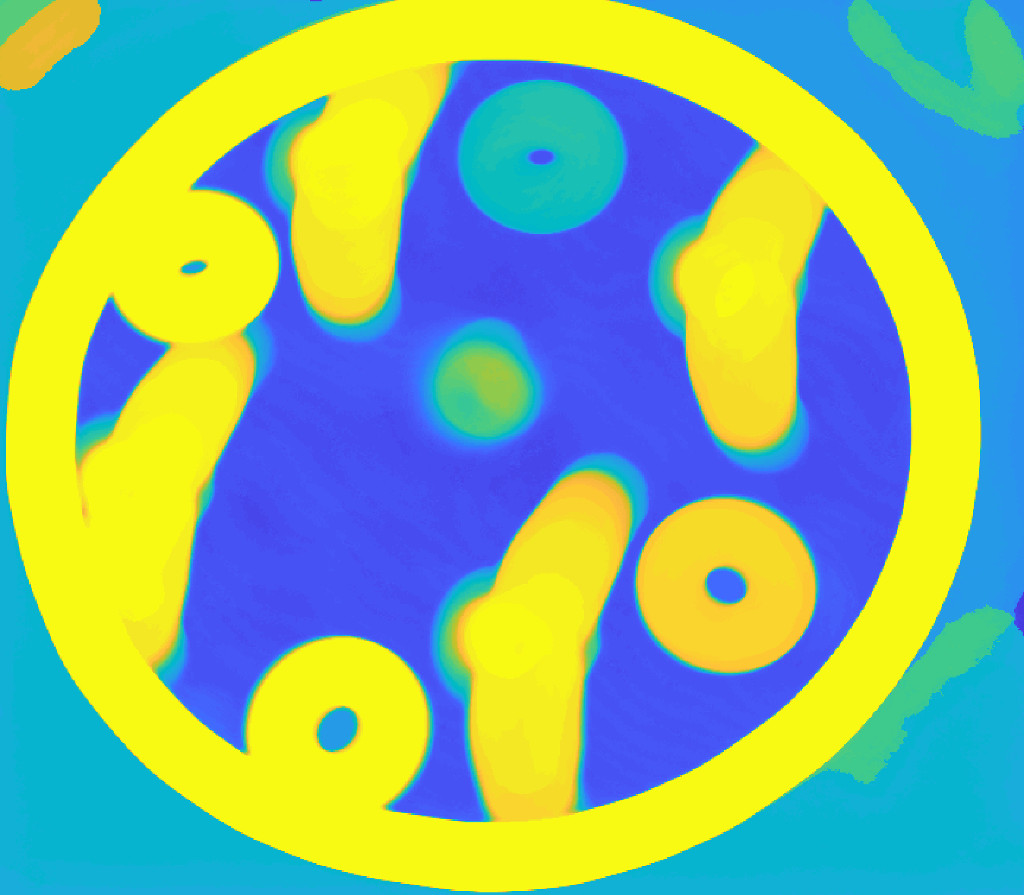}%
\label{fig:Moving_disk:5}}
\caption{(a) Image $f$ of the fixed object acquired with an exposure-time of 1/13~\textnormal{s}. The probe function $b$ is delineated in white. (d) Map of Asplund distances $Asp^{\protect \LP}_{b^{c},p} f^{c}$ of the complement of $f$. (b) Blurred image $f_{mov,bl}$ of the moving object acquired with the same exposure-time of 1/13~\textnormal{s}. (c) Disk detection in a dark image $f_{mov,dk}$ of the moving object acquired with a short exposure-time of 1/160~\textnormal{s}. The balls (delineated in white) are detected by finding the regional minima of its (e) map of LIP-additive Asplund distances $Asp^{\protect \LP}_{b^{c},p} f_{mov,dk}^{c}$. For each map, the tolerance parameter $p$ is set to 95\%.}
\label{fig:Moving_disk}
\end{figure}

The map of LIP-additive Asplund distances is also useful for detecting moving objects. In Fig. \ref{fig:Moving_disk}, a white disk with patterns is mounted on a turn table of a record player. The patterns include four small coloured disks and confounding shapes (i.e. eagles). First of all, an image $f$ with good contrasts is captured with an appropriate exposure-time, 1/13~s (Fig.~\ref{fig:Moving_disk:1}). In this image, a circular probe $b$ is selected inside a coloured disk. The record player is then started up at a speed of 45~tours/min and two images are captured. The first image $f_{mov,bl}$, which is acquired with the same exposure-time as the one of $f$, is correctly exposed but blurred (Fig.~\ref{fig:Moving_disk:2}). As it is blurred, it is useless to detect the coloured disks. A second image $f_{mov,dk}$ is acquired at a shorter exposure-time of 1/160~s. This second image is not blurred but darker than $f$ (Fig.~\ref{fig:Moving_disk:3}). The map of LIP-additive Asplund distances of its complement, $Asp^{\LP}_{b^{c},p} f_{mov,dk}^{c}$ (Fig.~\ref{fig:Moving_disk:5}), is useful for the detection of the coloured disks. Those are detected by finding the regional minima of the map with sufficient height using the h-minima transform \citep{Soille2003}. The regional minima with too large or too small area are also removed. By comparing the map of the moving disk image $Asp^{\LP}_{b^{c},p} f_{mov,dk}^{c}$ (Fig.~\ref{fig:Moving_disk:5}) to the map of the fixed disk image $Asp^{\LP}_{b^{c},p} f^{c}$ (Fig.~\ref{fig:Moving_disk:4}), one can notice that they both present similar shapes and amplitudes although the images $f$ and $f_{mov,dk}$ are captured with different conditions. In particular, the kinetics of the scene are different because the disk is fixed or turning, and the lighting conditions differ significantly because the exposure-times are varying by a factor 12. 

Fig. \ref{fig:Different_exposure_times_Christmas_balls} shows the robustness of the LIP-additive Asplund metrics under intensity variations caused by variable exposure-times. Fig. \ref{fig:Moving_disk} shows its efficiency to detect moving objects in dark images acquired with a small exposure-time which is necessary to capture an unblurred view of the object. Such cases occurs in many applications like medical images \citep{Noyel2017c} or industry \citep{Noyel2011,Noyel2013}. E.g. in industrial control the objects are often presented to the camera on a conveyor (linear, circular, etc.) whose speed varies with the production rate. 

\subsubsection{LIP-multiplicative metric: images acquired with a variable absorption of the medium}

The independence of the map of LIP-multiplicative Asplund distances under light variations due to different absorption (or opacity) of the object is verified with a montage we made. It is composed of a transparent tank. A paper with motives is stuck on one of its sides. On the opposite side, a camera is disposed to capture an image of the paper through a medium composed of the tank and its contents: a green colourant diluted into water. Three images $f_{\Gamma}$ (Fig.~\ref{fig:Different_absorption_Dilution:1}), $f_{3 \Gamma}$ (Fig.~\ref{fig:Different_absorption_Dilution:2}) and $f_{12 \Gamma}$ (Fig.~\ref{fig:Different_absorption_Dilution:3}) are acquired for increasing concentrations of the colourant: $\Gamma$, $3 \Gamma$ and $12 \Gamma$, where $\Gamma$ is the initial concentration of the colourant in the tank. One can notice that the image brightness decreases with the increase of the colourant concentration. A circular probe $b$ is manually selected in the brightest image $f_{\Gamma}$ in order to detect similar shapes in the two other images $f_{3 \Gamma}$ and $f_{12 \Gamma}$. A map of Asplund distances is computed on the complement of each image with the same probe $b$ and the same tolerance parameter $p=90\%$. One can notice that the three maps of images $Asp^{\protect \LT}_{b^{c},p} f_{\Gamma}^{c}$ (Fig.~\ref{fig:Different_absorption_Dilution:4}), $Asp^{\protect \LT}_{b^{c},p} f_{3 \Gamma}^{c}$ (Fig.~\ref{fig:Different_absorption_Dilution:5}) and $Asp^{\protect \LT}_{b^{c},p} f_{12 \Gamma}^{c}$ (Fig.~\ref{fig:Different_absorption_Dilution:6}) present similar amplitudes. They are segmented by the same thresholding technique -- at the $33^{rd}$ percentile -- and two area openings in order to remove the too small and too large regions. The selected regions are then dilated for display purpose. One can notice that all the disks are detected in the two darkest images $f_{3 \Gamma}$ (Fig.~\ref{fig:Different_absorption_Dilution:2}) and $f_{12 \Gamma}$ (Fig.~\ref{fig:Different_absorption_Dilution:3}) using the probe $b$ extracted in the bright image $f_{\Gamma}$. These results show the low sensitivity of the maps of LIP-multiplicative Asplund distances to light variations caused by different absorptions. Such a situation occurs in images acquired by transmission (e.g. X-rays, tomography, spectrophotometry, etc.) \citep{Jourlin2016}.\columnbreak


\begin{figure}[!t]
\centering
\subfloat[]{\includegraphics[width=0.33\columnwidth]{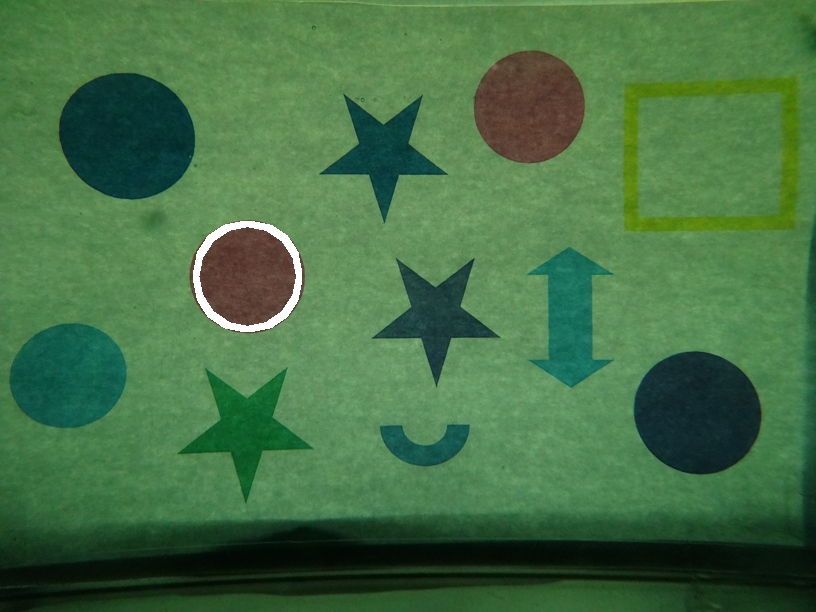}%
\label{fig:Different_absorption_Dilution:1}}
\hfil
\subfloat[]{\includegraphics[width=0.33\columnwidth]{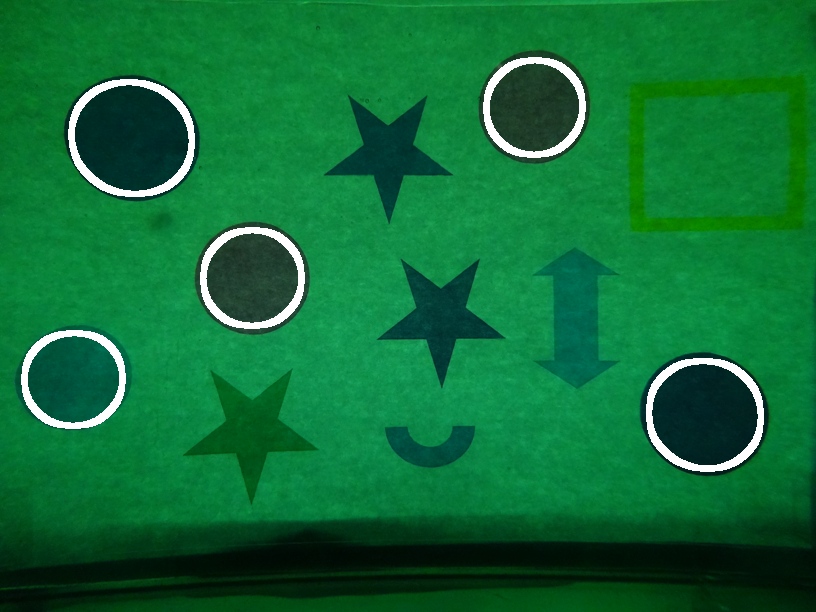}%
\label{fig:Different_absorption_Dilution:2}}
\hfil
\subfloat[]{\includegraphics[width=0.33\columnwidth]{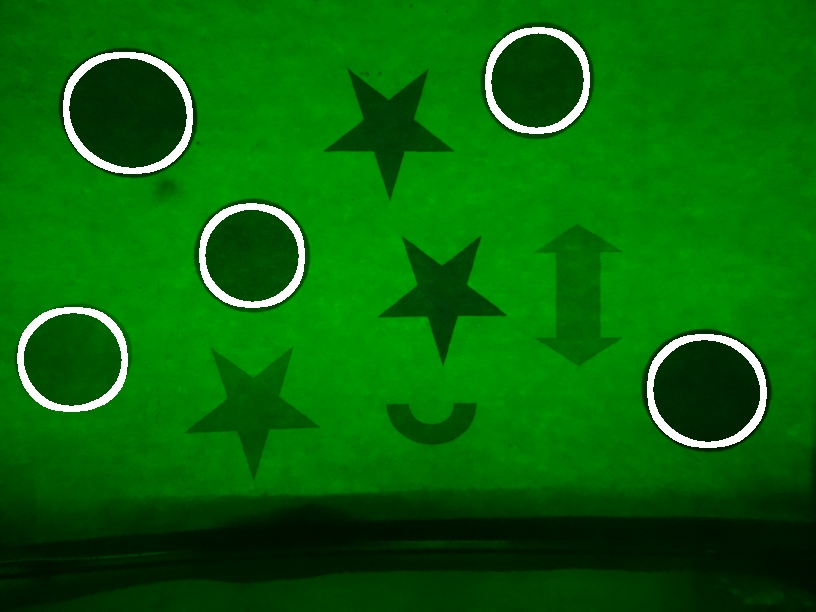}%
\label{fig:Different_absorption_Dilution:3}}
\hfil
\subfloat[]{\includegraphics[width=0.33\columnwidth]{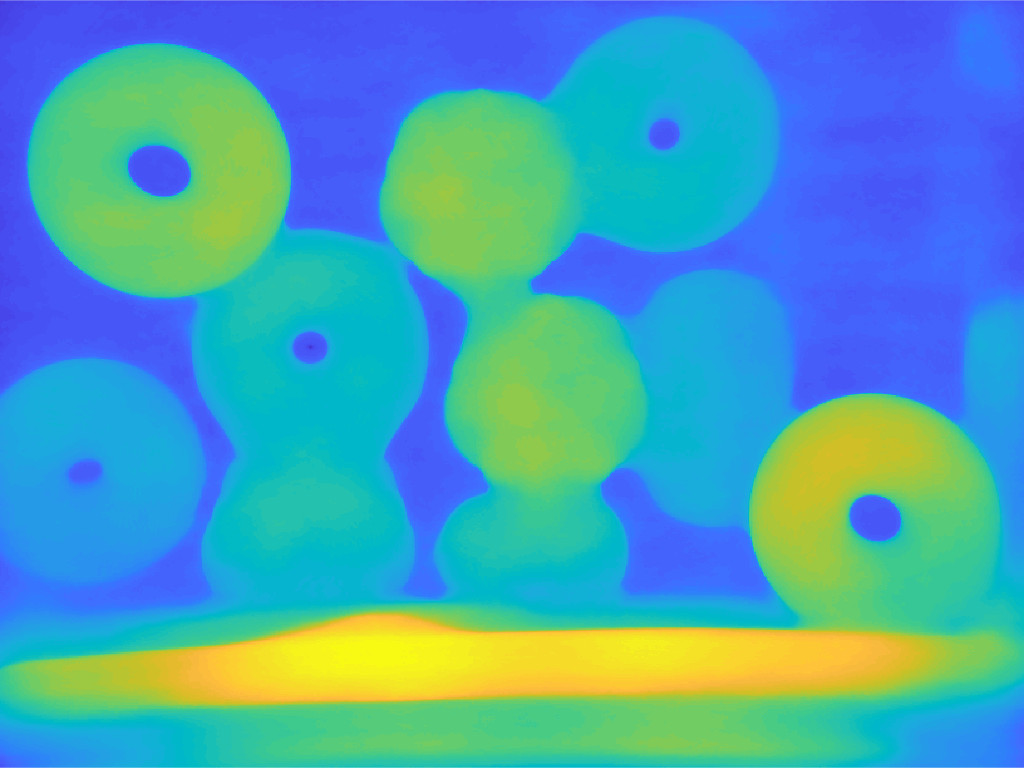}%
\label{fig:Different_absorption_Dilution:4}}
\hfil
\subfloat[]{\includegraphics[width=0.33\columnwidth]{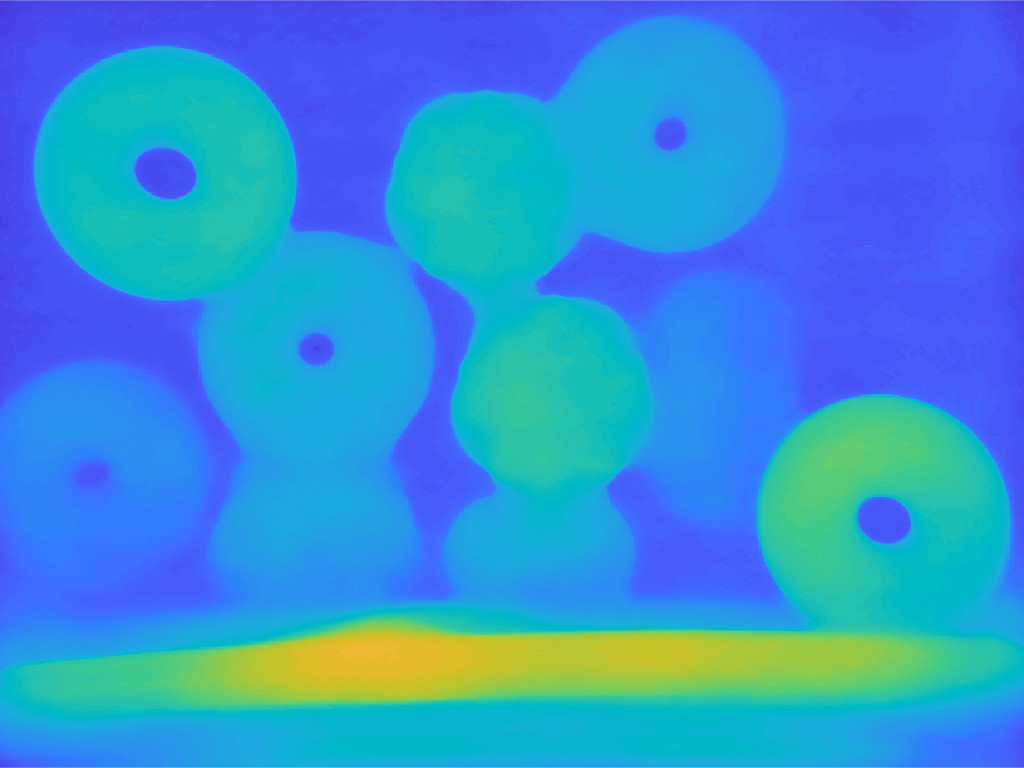}%
\label{fig:Different_absorption_Dilution:5}}
\hfil
\subfloat[]{\includegraphics[width=0.33\columnwidth]{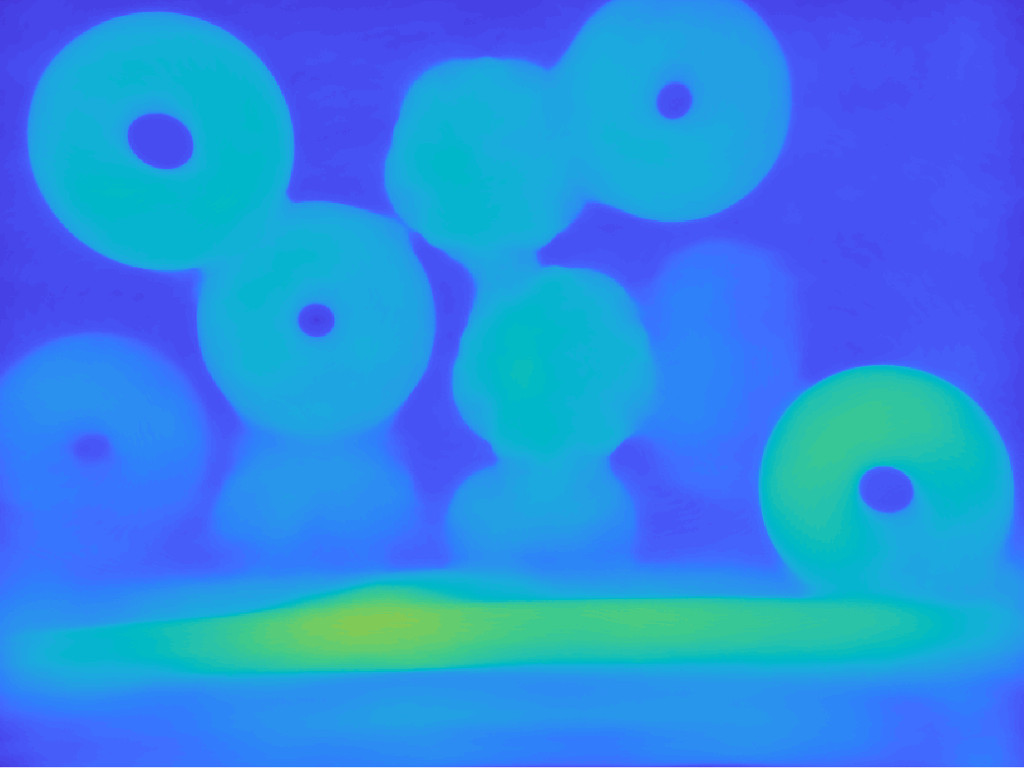}%
\label{fig:Different_absorption_Dilution:6}}
\caption{(a) Image $f_{\Gamma}$ acquired with a small concentration, $\Gamma$, of colourant. The probe function $b$ is delineated in white. (b) Ball detection in an image $f_{3 \Gamma}$ acquired with an intermediate concentration, $3 \Gamma$, of colourant. (c) Ball detection in an image $f_{12 \Gamma}$ acquired with a high concentration -- $12 \Gamma$ -- of colourant. (d) Map of LIP-multiplicative Asplund distances $Asp^{\protect \LT}_{b^{c},p} f_{\Gamma}^{c}$ of the  image $f_{\Gamma}$. (e) Map $Asp^{\protect \LT}_{b^{c},p} f_{3 \Gamma}^{c}$ of the image $f_{3 \Gamma}$. (f) Map $Asp^{\protect \LT}_{b^{c},p} f_{12 \Gamma}^{c}$ of the image $f_{12 \Gamma}$. The tolerance parameter $p$ is set to 90\%.}
\label{fig:Different_absorption_Dilution}
\end{figure}

\begin{remark}
In this section, probes with a circular invariance have been used to facilitate the presentation. However, the Asplund metrics are also efficient to detect non-circular objects with adapted probes. In addition, these metrics could be compared to the SIFT detector which is robust to lighting variations. The comparison has not been done in the sequel, because the robustness to lighting variations of the SIFT detector is not based on a physical law contrary to the Asplund metrics. However, \citet{Noyel2017c} have shown that enhancing the contrast of images improves registration methods based on SIFT points \citep{Lowe2004}. We could therefore develop a SIFT in the LIP framework. These findings will be the studied in a future paper.
\end{remark}


\section{Conclusion}
\label{sec:concl}

We have successfully presented a new framework of pattern matching robust to lighting variations between low-contrast and high-contrast images. It is composed of two metrics. Firstly, the LIP-multipli\-cative Asplund metric is robust to illumination changes due to variations of the object absorption or opacity. Secondly, the LIP-additive Asplund metric is robust to illumination changes caused by variations of the source intensity or of the camera exposure-time. Both metrics are respectively based on the multiplicative and the additive laws of the LIP model which give them strong optical properties. Both functional metrics are thereby theoretically \textit{insensitive to specific lighting variations}. They extend to images the property of \textit{insensitivity to object magnification} of the Asplund metric between binary shapes \citep{Asplund1960,Grunbaum1963}. After a presentation of the functional metrics and their properties, we have introduced robust to noise versions. We have demonstrated that the maps of Asplund distances between an image and a probe function are composed of Mathematical Morphology operations. Both maps of distances are especially related to the morphological operations of dilations and erosions for functions. Such a relation facilitates the programming of the maps of distances because these operations exist in numerous image processing libraries. The properties of both metrics have been then verified with simulated and real cases. Results have shown that the maps of LIP-multiplicative and LIP-additive Asplund distances are able to detect patterns in images acquired with different illuminations, caused by a stronger absorption of the object or by a shorter camera exposure-time, respectively. Importantly, the probe can be extracted in a highly contrasted image and the detection performed in a lowly contrasted image. Such properties pave the way to numerous applications where illumination variations are not controlled: e.g. in industry \citep{Noyel2013}, medicine \citep{Noyel2014,Noyel2017c}, traffic control \citep{Geiger2014}, safety and surveillance \citep{Foresti2005}, imaging of moving objects \citep{Noyel2011}, etc. In the future, these metrics will be extended to the analysis of colour and multivariate images starting from the preliminary ideas developed by \citet{Noyel2015,Noyel2016}. They will also be related to Logarithmic Mathematical Morphology \citep{Noyel2018}.



\appendix
\section{Appendix}
\label{sec:app}

The appendix is organised as follows. We will present the proofs of the propositions related to the LIP-multip\-licative Asplund metric (Prop. \ref{prop:dAs_Mult_0_M} and \ref{prop:expr_mapAsMult_morpho}) and to the LIP-additive metric (Prop. \ref{prop:gen_expr_lower_upper_maps_add}, \ref{prop:expr_maps_add_flat_se}, \ref{prop:property_lower_upper_maps_add} and \ref{prop:isomorph_LIPminus}).

\subsection{Proofs of propositions \ref{prop:dAs_Mult_0_M} and \ref{prop:expr_mapAsMult_morpho} related to the LIP-multiplicative Asplund metric}
\label{ssec:app:proofA}


\begin{proof}[Proof of proposition \ref{prop:dAs_Mult_0_M}, p. \pageref{prop:dAs_Mult_0_M}]
$c_1$ and $c_2$ can be expressed as: $c_1 = \bigvee_{x \in D}{ \{ \gamma(x) \} }$ and $c_2 = \bigwedge_{x \in D}{ \{ \gamma(x) \} }$ with the function $\gamma = f \LM g$, $\gamma \in \Fcurv_M= \left]-\infty,M\right[^D$. There always exists a constant $k= \bigwedge_{x \in D}{\{ \gamma(x) \}}$ such that $\gamma \LM k$ lies in $\I=[0,M[^D$ and is thus an image. Let us define $c_1^k = \bigvee_{x \in D}{ \{ \gamma(x) \LM k \} }$ and $c_2^k = \bigwedge_{x \in D}{ \{ \gamma(x) \LM k \} }$. $c_1^k$ and $c_2^k$ lie in $[0,M[$ and $c_1^k \geq c_2^k$. There is: \linebreak
$d_{As}^{\protect \LP}(f,g) = \bigvee_{x \in D}{ \{ \gamma(x) \} } \LM \bigwedge_{x \in D}{ \{ \gamma(x) \} } = $\linebreak$\bigvee_{x \in D}{ \{ \gamma(x) \LM k \} } \LM \bigwedge_{x \in D}{ \{ \gamma(x) \LM k \} } = c_1^k \LM c_2^k$.
Therefore, $d_{As}^{\LP}(f,g)$ lies in $[0,M[$ as the LIP-difference between  $c_1^k$ and $c_2^k \in \left[0,M\right[$, where $c_1^k \geq c_2^k$.
\label{proof:prop:dAs_Mult_0_M}
\end{proof}

\begin{proof}[Proof of proposition \ref{prop:expr_mapAsMult_morpho}, p. \pageref{prop:expr_mapAsMult_morpho} \citep{Noyel2017b}]
Let $f \in \overline{\I}$ be an image and $b~\in~\Tcurv^{D_b}$ be a probe. Using Eq.~\ref{eq:upper_map_2}, \ref{eq:dilation_funct_mult} and \ref{eq:rel_dil_mult_add} and knowing that $\tilde{f} \leq 0$, there is: 
$\forall x \in D$, \linebreak $\la_{b} f(x) = \bigvee_{-h \in D_b} \{ \tilde{f}(x-h) / \tilde{b}(-h) \}$. 
This leads to 
$\la_{b} f = (-\tilde{f}) \dot{\oplus} (-1/\tilde{\overline{b}}) = \exp{ ( \ln{(-\tilde{f})} \oplus (- \ln{(-\tilde{\overline{b}})) } ) } = \exp{ ( \hat{f} \oplus (- \hat{\overline{b}}) )}$.
\\Similarly, there is $\mu_{b} f = (-\tilde{f}) \dot{\ominus} (-\tilde{b}) = \exp{ [ \hat{f} \ominus \hat{b} ] } $.
The previous expressions of $\la_{b} f$ and $\mu_{b} f$ are used into Eq.~\ref{eq:map_AsMult_la_mu_general_se} to obtain: \linebreak
$Asp^{\LT}_{b}f	= \ln{ ( \exp{ [ \hat{f} \oplus (- \hat{\overline{b}}) ] } / \exp{ [ \hat{f} \ominus \hat{b} )} ])} = $\linebreak$ [ \hat{f} \oplus (- \hat{\overline{b}}) ] - [ \hat{f} \ominus \hat{b} ] = \delta_{-\hat{\overline{b}}} \hat{f} - \varepsilon_{\hat{b}} \hat{f}$.
\label{prop:dem:expr_maps_morpho}
\end{proof}

\subsection{Proofs of propositions \ref{prop:gen_expr_lower_upper_maps_add}, \ref{prop:expr_maps_add_flat_se}, \ref{prop:property_lower_upper_maps_add} and \ref{prop:isomorph_LIPminus} related to the LIP-additive Asplund metric}
\label{ssec:app:proofB}



\begin{proof}[Proof of proposition \ref{prop:gen_expr_lower_upper_maps_add}, p. \pageref{prop:gen_expr_lower_upper_maps_add}]
$\forall x \in D$, $\forall h \in D_b$, $\forall~c~\in~\Fcurv_M$, there is: \newline
$c(x) \LIPplus b(h) \geq f(x+h) \Leftrightarrow c(x) \geq f(x+h) \LIPminus b(h)$.

Eq.~\ref{eq:upper_map_add} becomes therefore:
$c_{1_{b}} f(x) =$\linebreak$ \inf \{c(x), c(x) \geq f(x+h) \LIPminus b(h), h \in D_b \}	= \bigvee \{ f(x+h) \LIPminus b(h) , h \in D_b \}$.
The last equality is due to the complete lattice structure. In a similar way, Eq.~\ref{eq:lower_map_add} becomes:
$c_{2_{b}} f(x) =$\linebreak$ \sup \{c(x), c(x) \leq f(x+h) \LIPminus b(h), h \in D_b \} =$\linebreak$ \bigwedge \{ f(x+h) \LIPminus b(h), h \in D_b \}$.
\end{proof}

%


\begin{proof}[Proof of proposition \ref{prop:expr_maps_add_flat_se}, p. \pageref{prop:expr_maps_add_flat_se}]
Let $b=b_0~\in~(\Tcurv)^{D_b}$ be a flat structuring element ($\forall x \in D_b$, $b(x) = b_0$). Knowing that $\LIPminus b_0$ preserves the order~$\leq$ (i.e. it is an increasing operator), Eq.~\ref{eq:upper_map_add_2} of $c_{1_{b}}$ and Eq.~\ref{eq:lower_map_add_2} of $c_{2_{b}}$ can be simplified: 
$\forall x \in D$,
$c_{1_{b_0}} f(x) = \bigvee \left\{ f(x+h) \LIPminus b_0,  h \in D_b \right\}	= \bigvee \left\{ f(x-h) ,  -h \in D_b \right\} \LIPminus b_0
							= \delta_{\overline{D}_b} f(x) \LIPminus b_0$.
Similarly, 
$c_{2_{b_0}} f(x) = \bigwedge \left\{ f(x+h) ,  h \in D_b \right\} \LIPminus b_0 = \varepsilon_{D_b} f(x) \LIPminus b_0$.
Eq.~\ref{eq:map_As_c1_c2_add_flat_se}, of the map of distances, is deduced from Eq.~\ref{eq:map_As_c1_c2_add} and the expressions of $c_{1_{b_0}} f$ and $c_{2_{b_0}} f$.
\end{proof}


\begin{proof}[Proof of proposition \ref{prop:property_lower_upper_maps_add}, p. \pageref{prop:property_lower_upper_maps_add}]
As $\LIPminus b(h)$ preserves the order $\leq$ (i.e. it is an increasing operator), there is:
$\forall f, g \in \Fcurvb_M$, $\forall x \in D$,\linebreak
$c_{1_{b}} ( f \vee g )(x) = \bigvee_{h \in D_b} \{ ( (f \vee g)(x+h) ) \LIPminus b(h) \}
=$\linebreak$ \bigvee_{h \in D_b} \{ ( f(x+h) \LIPminus b(h) ) \vee ( g(x+h) \LIPminus b(h) ) \}
= [\bigvee_{h \in D_b} \{ f(x+h) \LIPminus b(h)\} ] \vee [ \bigvee_{h \in D_b} \{ g(x+h) \LIPminus b(h)\} ]$\newline$
= c_{1_{b}} f(x) \vee c_{1_{b}} g(x)$.
\newline In addition, $\forall x \in D$, $c_{1_{b}} (O)(x) = c_{1_{b}}(f_{-\infty})(x) = \bigwedge_{h \in D_b} \{c(x), c(x) \geq (-\infty(x+h) \LIPminus b(h))\} =$\linebreak$ \bigwedge_{h \in D_b} \{c(x), c(x) \geq M (-\infty-b(h))/(M-b(h)) \} = -\infty = O(x)$, because $b(h)\in \left]-\infty,M\right[$. 
Therefore $c_{1_{b}}$ is a dilation (Def. \ref{pre:def_morpho_base}.2, p. \pageref{pre:def_morpho_base}). 

Similarly, $\forall f, g \in \Fcurvb_M$, 
$c_{2_{b}} ( f \wedge g )	= c_{2_{b}} ( f ) \wedge c_{2_{b}} ( g )$.\linebreak
In addition, $\forall x \in D$,
$c_{2_{b}} (I)(x) = c_{2_{b}}(f_M)(x)
							= \bigvee_{h \in D_b} \{c(x), c(x) \leq M(x+h) \LIPminus b(h) \}
							=$\linebreak$ \bigvee_{h \in D_b} \{c(x), c(x) \leq M (M - b(h))/(M-b(h)) \}
							= \bigvee_{h \in D_b} \{c(x), c(x) \leq M \}
							= M = I(x)$.
Therefore $c_{2_{b}}$ is an erosion (Def. \ref{pre:def_morpho_base}.1, p. \pageref{pre:def_morpho_base}).

\end{proof}


\begin{proof}[Proof of proposition \ref{prop:isomorph_LIPminus}, p. \pageref{prop:isomorph_LIPminus}]
Let $f, g~\in~\Fcurvb_M$ be two functions. There is:
$\xi(f \LP g) 	= -M\ln{[1 - (f \LP g)/M]}
								= -M\ln{[(1 - f/M)(1-g/M)]}$\linebreak$
								= -M\ln{(1 - f/M)} -M\ln{(1-g/M)}
								= \xi(f) + \xi(g)$, 
$\xi(\LM g) 		= -M\ln{[1 + g/(M-g)]} 
								= -M\ln{[M/(M-g)]}$\linebreak$
								= M\ln{(1 - g/M)} = -\xi(g)$, and
$\xi(f \LM g) 	=$\linebreak$ \xi(f \LP (\LM g)) = \xi(f) + \xi(\LM g) = \xi(f) - \xi(g)$.
\end{proof}


\section{Supplementary materials}
\label{sec:sup_mat}

The supplementary materials to this article include:
\begin{inparaenum}[i)]
\item a video abstract,
\item the proofs of propositions \ref{prop:gen_expr_lower_upper_maps}, \ref{prop:expr_maps_flat_se} and \ref{prop:property_lower_upper_maps},
\item the verification of the properties of the LIP-additive Asplund metric,
\item the proofs of the robust to noise metric invariances and
\item details about the illustration section. 
\end{inparaenum}

\section{Acknowledgments}
The authors are grateful to Dr Helen Boyle for her careful re-reading of the manuscript. 

%
%

\bibliography{refs}

\end{paper}
\end{document}